\documentclass[10pt, conference, compsocconf]{IEEEtran}

\usepackage{url}
\usepackage{cite}
\usepackage{amsmath,amssymb,amsfonts}
\usepackage{algorithmic}
\usepackage{graphicx}
\usepackage{textcomp}
\usepackage{paralist}
\usepackage{listings}
\usepackage{booktabs}
\usepackage{hhline}
\usepackage{framed,multirow}
\usepackage{soul}
\usepackage{cancel}
\usepackage{booktabs}
\usepackage{latexsym}
\usepackage{array}
\usepackage{stackengine}
\usepackage{relsize}
\usepackage[inline]{enumitem}
\usepackage[skip=2pt]{subcaption}
\usepackage[ruled,vlined]{algorithm2e}

\SetCommentSty{mycommfont}
\usepackage[dvipsnames]{xcolor}

\usepackage{url}
\usepackage{hyperref}

\hypersetup{
    colorlinks=true,
    linkcolor=blue,
    filecolor=magenta,      
    urlcolor=cyan,
}
\definecolor{newcolor}{rgb}{.8,.349,.1}

\newcommand{\ie}{\textit{i}.\textit{e}., }
\newcommand{\eg}{\textit{e}.\textit{g}., }
\newcommand{\cg}{\mathrm{CG}}
\newcommand{\tg}{\mathrm{TG}}

\newcommand{\dist}{\mathrm{dist}}
\newcommand{\gnn}{\mathrm{GNN}}
\newcommand{\cnn}{\mathrm{CNN}}

\newcommand{\troi}{\mathrm{TRoI}}
\newcommand{\hact}{\mathrm{HACT}}
\newcommand{\concat}{\mathrm{CONCAT}}
\newcommand{\lstm}{\mathrm{LSTM}}

\newcommand{\gin}{\mathrm{GIN}}
\newcommand{\pna}{\mathrm{PNA}}
\newcommand{\mlp}{\mathrm{MLP}}

\begin{document}

\title{Hierarchical Graph Representations in Digital Pathology}

\author{
\IEEEauthorblockN{Pushpak Pati*$^{1,2}$}
\and 
\IEEEauthorblockN{Guillaume Jaume*$^{1,3}$}
\and
\IEEEauthorblockN{Antonio Foncubierta-Rodríguez$^{1}$}
\and
\IEEEauthorblockN{Florinda Feroce$^{4}$}
\and
\IEEEauthorblockN{Anna Maria Anniciello$^{4}$}
\and
\IEEEauthorblockN{Giosue Scognamiglio$^{4}$}
\and
\IEEEauthorblockN{Nadia Brancati$^{5}$}
\and
\IEEEauthorblockN{Maryse Fiche$^{6}$}
\and
\IEEEauthorblockN{Estelle Dubruc$^{7}$}
\and
\IEEEauthorblockN{Daniel Riccio$^{5}$}
\and
\IEEEauthorblockN{Maurizio Di Bonito$^{4}$}
\and
\IEEEauthorblockN{Giuseppe De Pietro$^{5}$}
\and
\IEEEauthorblockN{Gerardo Botti$^{4}$}
\and
\IEEEauthorblockN{Jean-Philippe Thiran$^{3}$}
\and
\IEEEauthorblockN{Maria Frucci$^{5}$}
\and
\IEEEauthorblockN{Orcun Goksel$^{2}$}
\and
\IEEEauthorblockN{Maria Gabrani$^{1}$}
\and
\IEEEauthorblockA{$^1$IBM Research Zurich,\\ Switzerland}
\and
\IEEEauthorblockA{$^2$ETH Zurich,\\ Switzerland}
\and
\IEEEauthorblockA{$^3$EPFL,\\ Switzerland}
\and
\IEEEauthorblockA{$^4$IRCCS-Fondazione Pascale,\\ Italy}
\and
\IEEEauthorblockA{$^5$ICAR-CNR,\\ Italy}
\and
\IEEEauthorblockA{$^6$Aurigen,\\ Switzerland}
\and
\IEEEauthorblockA{$^7$CHUV,\\ Switzerland}
}

\maketitle

\begin{abstract}
Cancer diagnosis, prognosis, and therapy response predictions from tissue specimens highly depend on the phenotype and topological distribution of constituting histological entities.
Thus, adequate tissue representations for encoding histological entities is imperative for computer aided cancer patient care.
To this end, several approaches have leveraged cell-graphs that encode cell morphology and cell organization to denote the tissue information.
These allow for utilizing graph theory and machine learning to map tissue representations to tissue functionality to help quantify their relationship. Though cellular information is crucial, it is incomplete alone to comprehensively characterize complex tissue structure. 
We herein treat the tissue as a hierarchical composition of multiple types of histological entities from fine to coarse level, capturing multivariate tissue information at multiple levels. 
We propose a novel multi-level hierarchical entity-graph representation of tissue specimens to model hierarchical compositions that encode histological entities as well as their intra- and inter-entity level interactions. 
Subsequently, a hierarchical graph neural network is proposed to operate on the hierarchical entity-graph representation to map the tissue structure to tissue functionality. 
Specifically, for input histology images we utilize well-defined cells and tissue regions to build HierArchical Cell-to-Tissue ($\hact$) graph representations, and devise $\hact$-Net, a message passing graph neural network, to classify such $\hact$ representations. 
As part of this work, we introduce the BReAst Carcinoma Subtyping (BRACS) dataset, a large cohort of Haematoxylin \& Eosin stained breast tumor regions-of-interest, to evaluate and benchmark our proposed methodology against pathologists and state-of-the-art computer-aided diagnostic approaches. Through comparative assessment and ablation studies, our proposed method is demonstrated to yield superior classification results compared to alternative methods as well as individual pathologists.
\end{abstract}


\IEEEpeerreviewmaketitle

\section{Introduction}
\label{sec:introduction}

Breast cancer is the most commonly diagnosed cancer and registers the highest number of deaths for women with cancer~\cite{sung21}. A study by~\cite{allemani15} exhibits that intensive early diagnostic activities have improved 5-year survival to 85\% during 2005–09 for breast cancer patients. Early diagnosis of cancer, primarily through manual inspection of histology slides, enables the acute assessment of risk and facilitates an optimal treatment plan. Though the diagnostic criteria for breast cancer are established, the continuum of histologic features phenotyped across the diagnostic spectrum prevents the distinct demarcation. Thus, manual inspection is tedious and time-consuming with significant intra- and inter-observer variability~\cite{gomes14, elmore15}. Increasing incidence rate of breast cancer cases per year~\cite{siegel20} and the complications in manual diagnosis demand for automated computed-aided diagnostic tools.

Whole-slide scanning systems empowered rapid digitization of pathology slides into high-resolution whole-slide images (WSIs), thereby profoundly transforming pathologists' daily practice~\cite{mukhopadhyay17}. Further, they enabled computed aided diagnostics to leverage artificial intelligence~\cite{litjens17, deng20}, especially deep learning, to address various digital pathology tasks in breast diagnosis~\cite{ibrahim20}, such as
nuclei segmentation~\cite{kumar17,graham19}, 
nuclei classification~\cite{pati20-media,verma20},
gland segmentation~\cite{graham19-mild,binder19},
tumor detection~\cite{aresta19,bejnordi17-jama,pati18}, and
tumor staging~\cite{aresta19,mercan19}.
Deep learning techniques primarily employ Convolutional Neural Networks ($\cnn$)~\cite{madabhushi16,parwani19} to process pathology images in a patch-wise manner. $\cnn$s extract representative patterns from patches and aggregate patch representations to perform image-level tasks.
However, patch-wise processing suffers from the trade-off between the resolution of operation and the acquisition of adequate context~\cite{bejnordi17, sirinukunwattana18}.
Operating at a higher resolution captures local cellular information but limits the field-of-view due to computational burden, thus limiting access to global information that would capture the immediate tissue microenvironment. In contrast, operating at a lower resolution hinders resolvability of cells and access to cellular features.~\cite{bejnordi17, sirinukunwattana18, tellez20} have proposed $\cnn$ methods to address such trade-off by leveraging visual context, however $\cnn$s, which operate on fix-sized input patches, are confined to a fixed field-of-view and are restricted for diagnosis from incorporating information from varying spatial distances.
Further, pixel-based processing in $\cnn$s disregards the notion of pathologically meaningful entities~\cite{hagele20}, such as cells, glands, and tissue types. Such absence of a notion for relevant entities severely limits the interpretability of $\cnn$s and any utilization of well-established entity-level prior pathological knowledge in $\cnn$-based digital pathology frameworks. Additionally, $\cnn$s disregard the structural composition of tissue, where fine entities hierarchically constitute to form coarser entities, such as, epithelial cells organize to form epithelium, which further constitutes to form glands. Such hierarchical structure is relevant both for diagnostics and for interpretation.

In this paper, we address the aforementioned limitations by shifting the analytical paradigm from pixel to entity-based processing. In an entity paradigm, a histology image is described as an entity-graph, where nodes and edges of a graph denote biological entities and inter-entity interactions, respectively. An entity-graph can be customized in various aspect, \eg in terms of the type of entity set, entity attributes, and graph topology, by incorporating any task-specific prior pathological knowledge. Thus, the graph representation enables pathology-specific interpretability and human-machine co-learning. In addition, the graph representation is memory efficient compared to pixelated images and can seamlessly describe a large tissue region.
\cite{demir04} first introduced cell-graphs using cells as the entity type. Though a cell-graph efficiently encodes the cell microenvironment, it cannot extensively capture the tissue microenvironment, \ie the distribution of tissue regions such as necrosis, stroma, epithelium, etc. Similarly, a tissue-graph comprising of the set of tissue regions cannot depict the cell microenvironment. Therefore, an entity-graph representation using a single type of entity set is insufficient to comprehensively describe the tissue structure.
To address the limitation, we propose a multi-level entity-graph representation, \ie HierArchical Cell-to-Tissue ($\hact$), consisting of multiple types of entity sets, \ie cells and tissue regions, to encode both cell and tissue microenvironment. The multiset of entities is inherently coupled depicting tissue composition at multiple scales. The $\hact$ graph encodes individual entity attributes and intra- and inter-entity relationships to hierarchically describe a histology image.
Upon the graph construction, a graph neural network ($\gnn$), a deep learning technique operating on graph-structured data, processes the entity-graph to perform image analysis.
Specifically, we introduce a hierarchical $\gnn$, HierArchical Cell-to-Tissue Network ($\hact$-Net), to sequentially operate on $\hact$ graph, from fine-level to coarse-level, to provide a fixed dimensional embedding for the image. The embedding encodes morphological and topological distribution of the multiset of entities in the tissue.
Interestingly, our proposed methodology resembles the tissue diagnostic procedure in clinical practice, where a pathologist hierarchically analyzes a tissue. 

We propose a methodology that consists of $\hact$ graph construction and $\hact$-Net based histology image analysis. We characterize breast tumor regions-of-interest ($\troi$s) to evaluate our methodology. A preliminary version of this work was presented as~\cite{pati20}. 
Our substantial extensions herein include 
1)~an improved $\hact$ representation and $\hact$-Net architecture, 
2)~a larger evaluation dataset (twice the earlier size), 
3)~detailed ablation studies and evaluation on public data, and 
4)~a benchmark comparison against independent pathologists.
Specifically, the major contributions of this paper are:
\begin{itemize}
    \item A novel hierarchical entity-graph representation ($\hact$) and hierarchical learning ($\hact$-Net) methodology for analyzing histology images;
    \item Introducing a public dataset, BReAst Carcinoma Subtyping (BRACS\footnote{The BRACS dataset for breast cancer subtyping will be shortly released}), a large cohort of breast $\troi$s annotated with seven breast cancer subtypes. BRACS includes challenging atypical cases and a wide variety of $\troi$s representing a more realistic breast cancer analysis;
    \item An evaluation of our proposed methodology on the BRACS dataset by comparing with three independent pathologists, where an extensive assessment demonstrates our classification performance outperforming several recent $\cnn$ and $\gnn$ approaches for cancer subtyping, while being comparable to pathologists on per-class and aggregated classification tasks.
\end{itemize}


\section{Related work}
\label{sec:related_work}

\textbf{Tumor subtyping in digital pathology:}
Several deep learning algorithms have been proposed to categorize pathology images or WSIs into cancer subtypes~\cite{komura18, srinidhi19, deng20, spanhol16, araujo17, aresta19}. For this task, most algorithms in the literature employ $\cnn$s in a patch-wise manner: 
In~\cite{araujo17, bardou18, roy19, mercan19} $\cnn$s are utilized to categorize breast cancer pathology images into various subtypes. These methodologies employ single stream patch-wise approaches to capture local patch-level context, unify patch-level information via several aggregation strategies, and classify the image using aggregated information. 
However, single-stream approaches do not capture adequate context from neighboring tissue microenvironment to aptly classify a patch. 
In~\cite{sirinukunwattana18} this issue is addressed by including multi-scale information from concentric patches across different magnifications.
\cite{tellez20} propose a neural image compression methodology, where WSIs are compressed using a neural network trained in an unsupervised fashion, followed by a $\cnn$ trained on the compressed image representations to predict image labels. 
\cite{shaban20} include an attention module with an auxiliary task to improve neural image compression for classifying histology images.
\cite{yan20} propose a hybrid convolutional and recurrent deep neural network to utilize spatial correlations among patches for pathological image classification.
\cite{bejnordi17} propose a stacked $\cnn$ architecture to capture large contexts and perform end-to-end processing of large-sized histology images.
\cite{pinckaers20} propose a streaming $\cnn$ to accommodate multi-megapixel images. 
\cite{campanella19} utilize a multiple instance learning approach to process whole-slide images in an end-to-end manner, which is extended by~\cite{lu20} to automatically identify sub-regions of high diagnostic value through an attention mechanism.
Whereas the aforementioned methodologies employ different strategies to capture useful context information, they all still operate on a square and fix-sized input image. 
However, actual $\troi$s can be of highly varying dimensions and shapes depending on cancer subtype and the site of tissue extraction. 
Our proposed entity-graph methodology can acquire both local and global context from arbitrary-sized $\troi$s to address the aforementioned limitations.

\textbf{Graphs in digital pathology:}
Entity-graph based tissue representations can effectively describe the phenotypical and structural properties of tissues by incorporating morphology, topology, and interactions among biologically relevant tissue entities. 
Using cells as entities,~\cite{demir04} introduced a cell-graph ($\cg$) representation of tissues, where cell morphology can be embedded in the nodes via hand-crafted~\cite{demir04,zhou19,pati20} or deep-learning based features~\cite{chen20}.
The graph topology is often heuristically defined, \eg using k-Nearest Neighbors, probabilistic modeling, or a Waxman model~\cite{sharma15}. 
Subsequently, a $\cg$ is typically processed by classical machine learning techniques~\cite{sharma16,sharma17} or $\gnn$s~\cite{zhou19,pati20,chen20,anand20} for mapping to tissue function. 
Recently, improved graph representations using patches~\cite{aygunes20} and tissue regions~\cite{pati20} have been proposed to enhance the tissue structure-function mapping. 
Other graph-based applications in computational pathology include cellular community detection~\cite{javed20} and WSI classification~\cite{zhao20,adnan20}.
Notably, entity-graph representations consist of biological entities which the pathologists can readily relate to. Thus, graph-based analysis allows for incorporating pathologically-defined, task-specific, entity-level prior knowledge in constructing ``meaningful'' tissue representations. 
This implicitly enables and facilitates the study of \emph{interpretability} and \emph{explainability} of such graph-based networks by pathologists. To this end,~\cite{zhou19} analyze the clustering of nodes in a $\cg$ in order to group cells according to their appearance and tissue types.~\cite{jaume20} introduce a post-hoc graph-pruning explainer to identify decisive cells and interactions.~\cite{sureka20} employ a robust spatial filtering that utilizes an attention-based $\gnn$ and node occlusion to highlight cell contributions.~\cite{jaume21} propose quantitative metrics based on pathologically measurable cellular properties in order to characterize several graph explainability algorithms in $\cg$ analysis.



\section{Preliminaries}

\subsection{Notation}
We define an attributed, undirected entity-graph $G := (V, E, H)$ as a set of nodes $V$, edges $E$, and node features $H$. Each node $v \in V$ is represented by a feature vector $h(v) \in \mathbb{R}^d$, thus, $H \in \mathbb{R}^{|V|\, \times\, d}$. $d$ denotes the number of features per node, and $|\,.\,|$ denotes set cardinality. An edge between two nodes $u, v \in V$ is denoted as $e_{uv}$. The graph topology is described by a symmetric adjacency matrix $A \in \mathbb{R}^{|V|\, \times\, |V|}$, where $A_{u,v} = 1$ if $e_{uv} \in E$. The neighborhood of a node $v \in V$ is denoted as $\mathcal{N}(v) := \{u \in V \; | \; v \in V, \; e_{uv} \in E \;\}$.

\subsection{Graph Neural Networks}
$\gnn$~\cite{deferrard16, kipf17, xu19, hamilton17, velikovic18} defines a class of neural networks that extend the concept of convolution to operate on graph-structured data. 
In this work, we employ a $\gnn$ based on message-passing. 
In message-passing $\gnn$s~\cite{gilmer17}, node features $h(v), \, \forall v \in V$ are iteratively updated in a two-step procedure, i)~\textit{AGGREGATE}, and ii)~\textit{UPDATE}.
In the \textit{AGGREGATE} step for node $v$, the features of neighboring nodes $\mathcal{N}(v)$ are aggregated into a single feature representation. 
In the \textit{UPDATE} step, the features of node $v$ is updated by using the current node features and the aggregated representation from the \textit{AGGREGATE} step.
A series of $T$ such iterations, in the form of $\gnn$ layers, are employed to obtain updated node features $\forall \, v \in V$, incorporating information within upto $T$-hops from each node.
Finally, the node features $h^T(v)$ are pooled in the \textit{READOUT} step to build a fix-sized graph-level embedding $h_G$. 
\textit{AGGREGATE}, \textit{UPDATE}, and \textit{READOUT} operations must be differentiable to allow back-propagation in $\gnn$ training. Additionally, \textit{AGGREGATE} and \textit{READOUT} operations must be permutation-invariant such that the aggregated representation is invariant to node ordering.
Formally, the three steps are presented as 
\begin{equation}
\label{eqn:gnn_preliminary}
\begin{split}
a^{\, t \,+\,1}(v) & = \mathit{AGGREGATE}\, (\, \{\, h^{\,t}(u) : u \in \mathcal{N}(v) \, \} \,) \\
h^{\, t \,+\,1}(v) & = \mathit{UPDATE}\,(\,h^{\,t}(v), \, a^{\, t \,+\,1}(v)\,) \\
h_{\,G} & = \mathit{READOUT}\,(\,\{\,h^{\,T}(v) : v \in V\,\}\,)
\end{split}
\end{equation}

\begin{figure}
\centering
\subfloat[][\centering $\gin$]{{\includegraphics[width=0.36\linewidth]{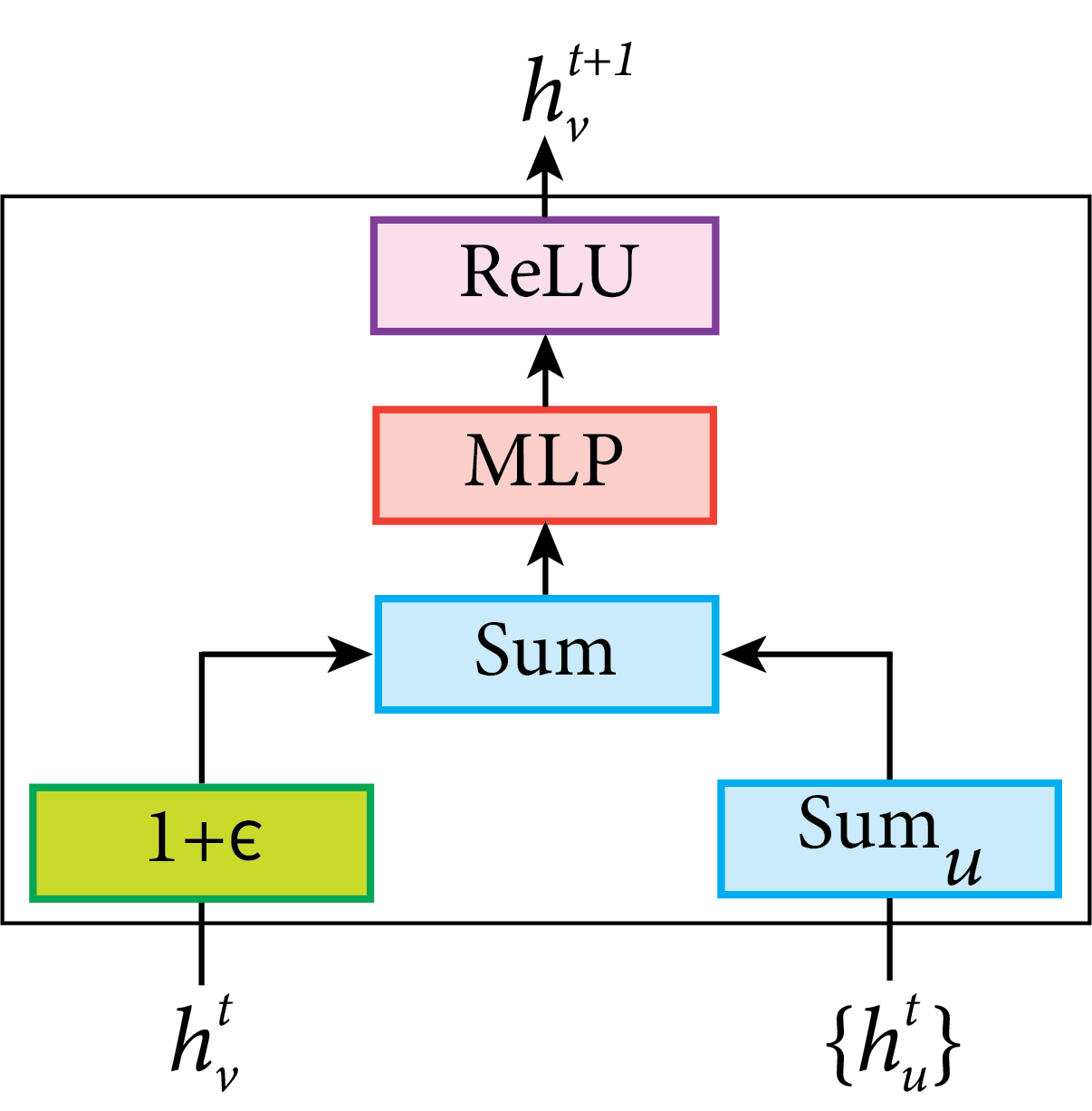}}}
\quad
\subfloat[][\centering $\pna$]{{\includegraphics[width=0.59\linewidth]{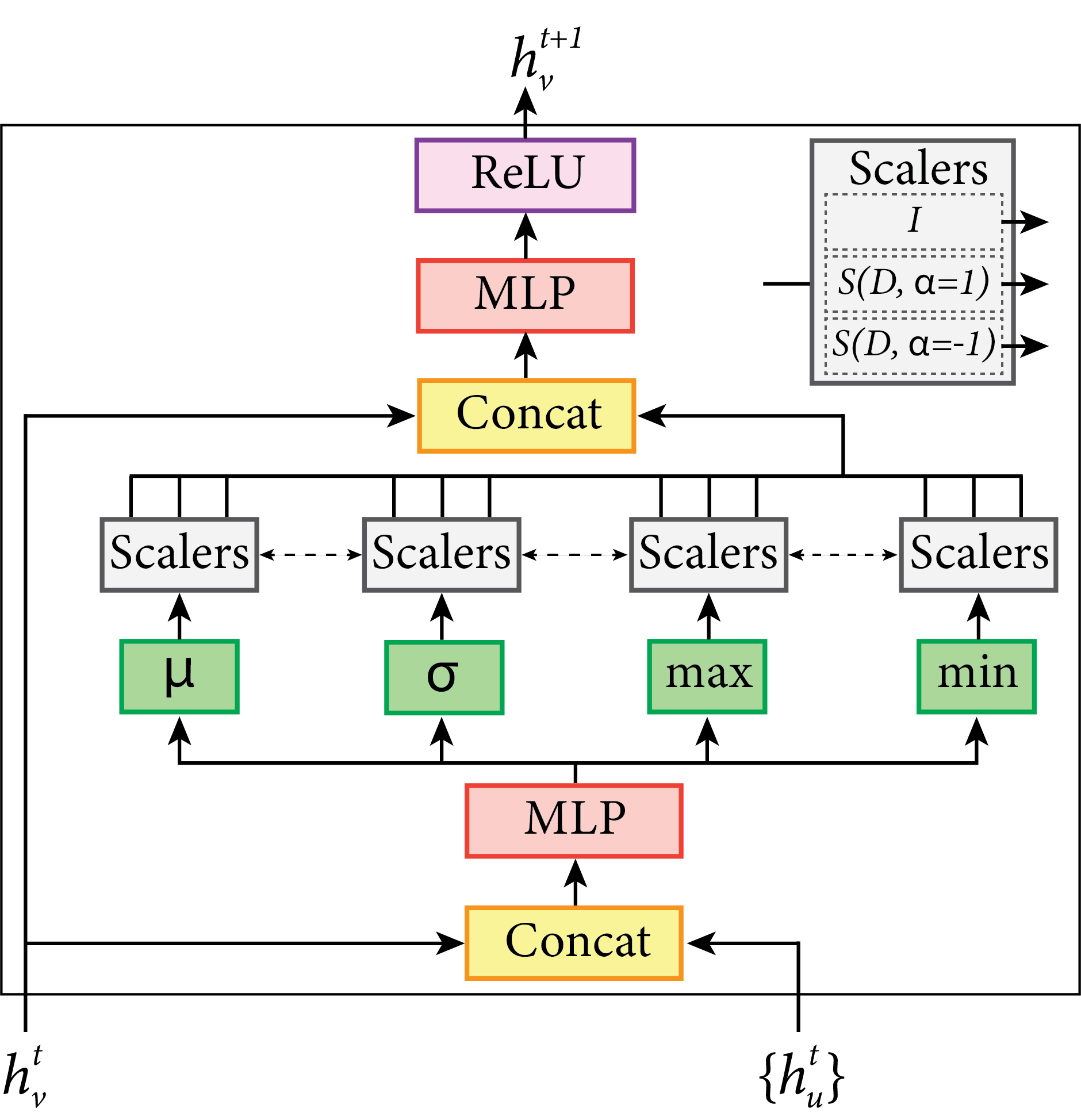}}}
\caption{Overview of $\gin$ and $\pna$ layers, where $h_v^t$ and $\{h_u^t\}$ denote the  representation of, respectively, node $v$ and its neighbors at layer $t$. $\gin$ uses sum as the \textit{AGGREGATE} function, followed by a sum and multi-layer perceptron ($\mlp$) for the \textit{UPDATE} function. $\pna$ uses a set of aggregators (element-wise mean, standard deviation, maximum, and minimum) followed by degree-scalers (identity, amplifier, and dampener) as the \textit{AGGREGATE} function. The \textit{UPDATE} function consists of a concatenation followed by an $\mlp$.}
\label{fig:gin_pna}
\end{figure}

An important aspect of designing a $\gnn$ is the characterization of its expressive power. A $\gnn$ has a strong expressive power if it can map two non-isomorphic graphs to two unique graph embeddings, thus imparting an injective mapping between the graph and embedding spaces. 
A line-of-research exploring the expressive power of $\gnn$s~\cite{xu19, morris2018, jaume19} highlight the connection between iterative message passing procedure of $\gnn$ and the popular Weisfeiler-Lehman (WL)~\cite{weisfeiler1968}  test for graph isomorphism. It is established that architectures such as the Graph Isomorphism Network (GIN)~\cite{xu19} can perform as well as the 1-dimensional WL test for \emph{countable} node feature spaces, \ie when the node features are discrete. An example of graph with discrete node features can be the study of molecule design, where the nodes represent atoms that are discrete in nature. 
Recent studies show that in case of \emph{continuous} node features, \eg $\cnn$-based node embeddings, the use of multiple permutation-invariant aggregators, such as sum, max, and mean, can help in building expressive $\gnn$s~\cite{dehmany19, corso20}. To this end,~\cite{corso20} proposed the Principal Neighbourhood Aggregation ($\pna$) network by using a combination of aggregators followed by \emph{degree-scalers}. The series of aggregators replace the sum operation in $\gin$ while the degree-scalers can amplify or dampen neighboring aggregated-messages based to the degree of a node. 
Illustrations of $\gin$ and $\pna$ architectures are shown in Figure~\ref{fig:gin_pna}.

\begin{figure*}[!t]
\centering
\centerline{\includegraphics[width=0.98\textwidth]{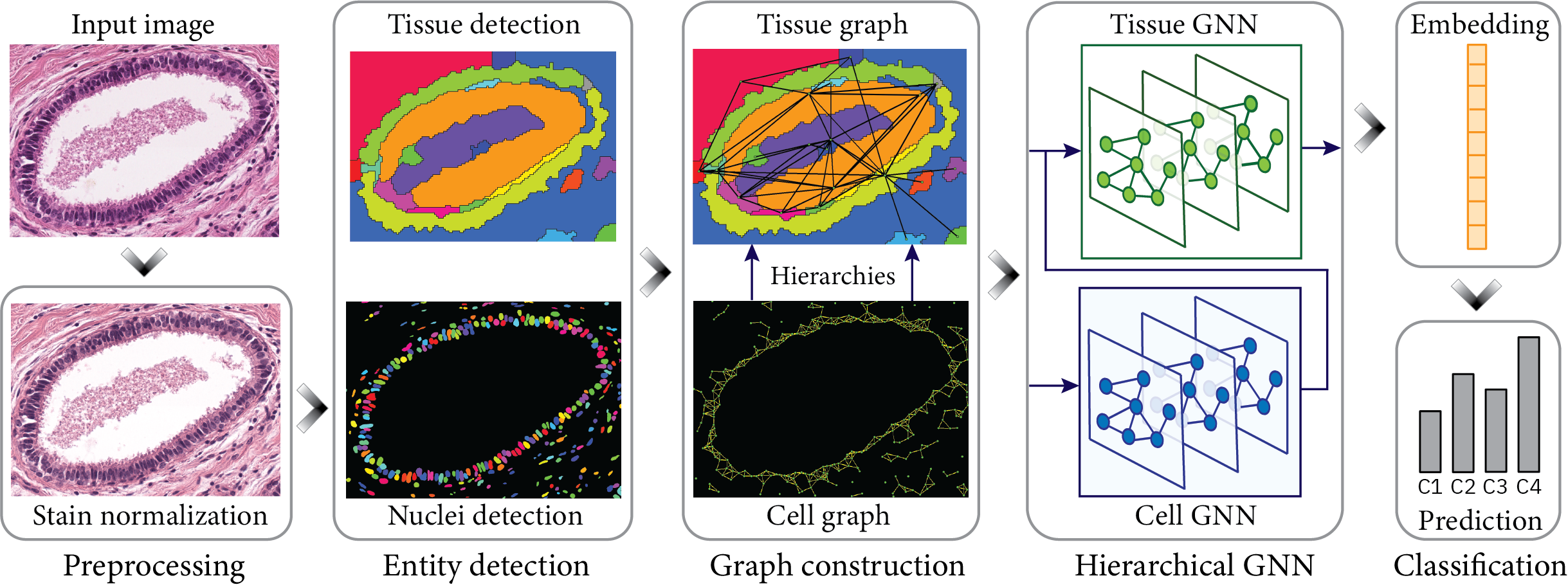}}
\caption{Overview of the proposed hierarchical entity-graph based tissue analysis methodology. Following some pre-processing, a hierarchical entity-graph representation of a tissue is constructed, and it is processed via a hierarchical graph neural network to learn the mapping from tissue compositions to respective tissue categories. (Figure is best viewed in color)}
\label{fig:framework}
\end{figure*}

\section{Methodology}
\label{sec:methodology}
In this section, we detail our proposed methodology for hierarchical tissue analysis, as illustrated in Figure~\ref{fig:framework}. For an input Hematoxylin and Eosin (H\&E) stained histology $\troi$ image, first, we apply some pre-processing to standardize the input. Then, we identify pathologically relevant entities and construct a $\hact$ graph representation of the $\troi$ by encoding the morphological and topological information of such entities. Finally, $\hact$-Net, a hierarchical $\gnn$, is employed to map the $\hact$ graph to a corresponding tissue class, \eg a cancer subtype.

\subsection{Pre-processing}
\label{sec:preprocessing}
H\&E stained tissue specimens exhibit appearance variability due to various reasons, such as different specimen preparation techniques, staining protocols (\eg temperature of the solutions used), fixation characteristics, and imaging device characteristics. Such variability in appearance adversely impacts any computational methods for downstream diagnosis~\cite{veta14, tellez19}. To alleviate the appearance variability, we employ the unsupervised, reference-free stain normalization algorithm proposed by~\cite{macenko09}. 
The algorithm is based on the principle that RGB color of each pixel is a linear combination of two unknown stain vectors, Hematoxylin and Eosin, that need to be estimated. First, the algorithm estimates the stain vectors of a $\troi$ by using a Singular Value Decomposition of the non-background pixels.
Second, the algorithm applies a correction to account for the intensity variations due to noise. The algorithm does not involve any intermediate step that may require model parameter tuning or training. 
Specifically, for stain normalization we employ the scalable and fast pipeline proposed by~\cite{stanisavljevic18}.

\subsection{Graph representation}
\label{sec:graph_representation}
Stain normalized $\troi$s are next processed to identify entities relevant to constructing a hierarchical entity-graph representation.
Given a histological task, different choices of entities in histology images can be selected as the relevant biological structures.
In this work, we consider nuclei and tissue regions as the relevant entities, such that our $\hact$ graph representations consist of three components: 1)~a low-level \emph{cell-graph}, capturing cell morphology and interactions, 2)~a high-level \emph{tissue-graph}, capturing morphology and spatial distribution of tissue regions, and 3)~cells-to-tissue hierarchies, encoding, \eg the relative spatial distribution of cells with respect to the tissue distribution. The details of the components are presented in the following subsections. 

\begin{figure*}
\centering
\centerline{\includegraphics[width=0.98\textwidth]{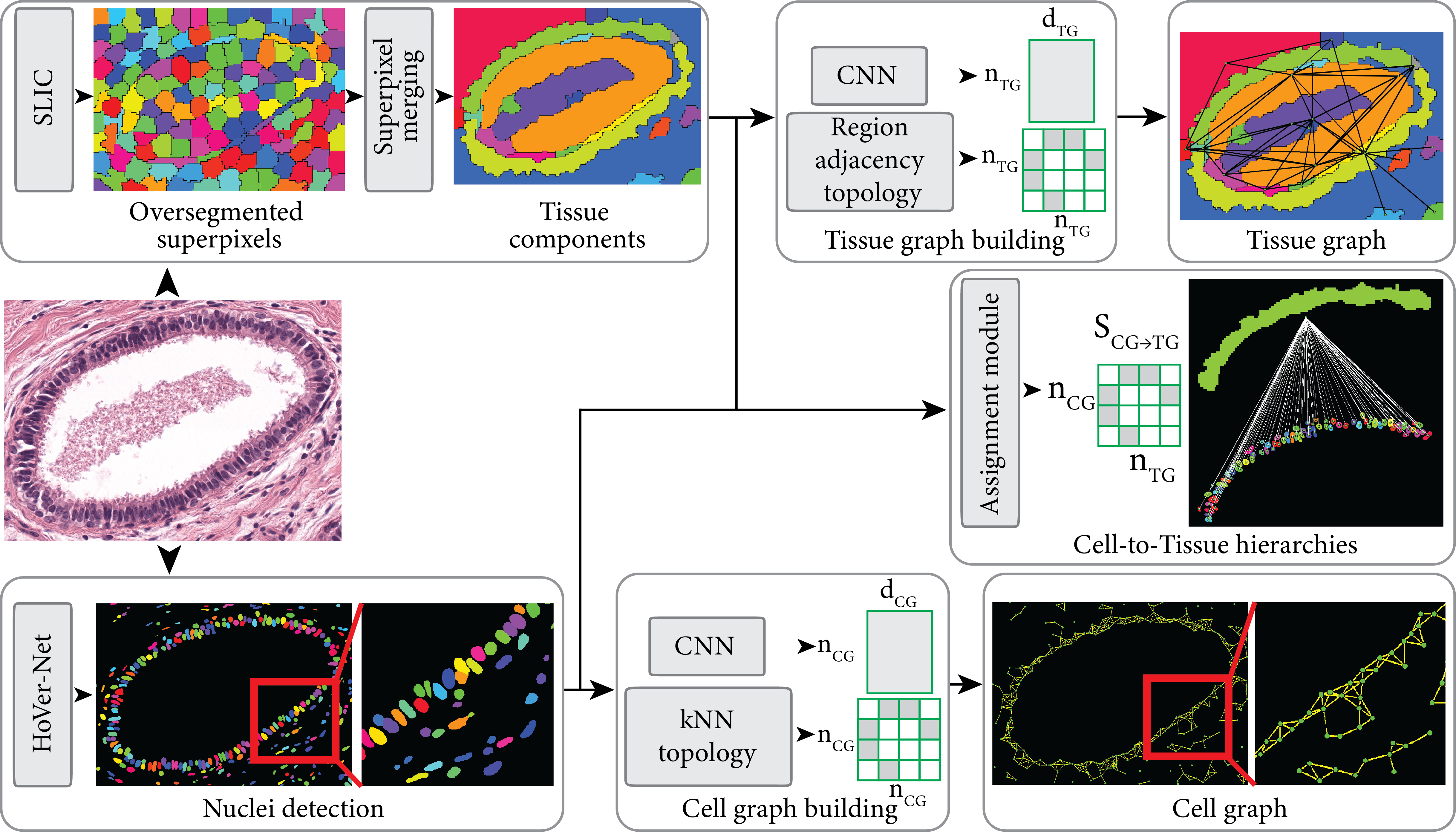}}
\caption{Overview of hierarchical cell-to-tissue ($\hact$) graph construction for a $\troi$. Our $\hact$ graph representation consists of a cell-graph, a tissue-graph, and cell-to-tissue hierarchies, while encoding the phenotypical and topological distributions of tissue entities to describe the cell and tissue microenvironments. (Figure is best viewed in color)}
\label{fig:graph_generation}
\end{figure*}

\subsubsection{Cell-graph representation}
\label{sec:cell_graph_representation}
A cell-graph ($\cg$) characterizes low-level cell information, where the nodes represent cells encoding cell morphology, and the edges encode cellular interactions depicting cell topology. A $\cg$ is constructed in three steps, i)~nuclei detection, ii)~nuclei feature extraction, and iii)~topology configuration. The steps are demonstrated in Figure~\ref{fig:graph_generation}.

Precise nuclei detection facilitates reliable $\cg$ representation. To this end, we use HoVer-Net, a nuclei segmentation network proposed by~\cite{graham19}, pre-trained on MoNuSeg dataset by~\cite{kumar17}. HoVer-Net leverages the instance-rich information encoded within the vertical and horizontal distances of nuclear pixels to their centers of mass. These distances are used to separate clustered nuclei, resulting in an accurate segmentation, particularly in areas with overlapping instances. 
The centroids of the segmented instances are used as the spatial coordinates of nodes in $\cg$.

Following nuclei detection, morphological features are extracted by processing patches of size $h \times w$ centered around nuclei centroids via ResNet~\cite{he16} architecture pre-trained on ImageNet dataset~\cite{deng09}. Spatial features of the nuclei are extracted as the spatial coordinates of the nuclei, normalized by the $\troi$ dimensions. Morphological and spatial features together constitute the nuclei features, which are colocated for all nodes as the node-feature matrix $H_{\cg} \in \mathbb{R}^{\;|V_{\cg}| \, \times \, d_{\cg}}$.

For the $\cg$ topology $E_{\cg}$, we utilize the fact that spatially close cells have stronger interactions with distant cells having weaker cellular interactions~\cite{francis97}.
Accordingly, we connect nearby cells with edges to model their interactions.
To this end, we use the k-Nearest Neighbors (kNN) algorithm to build an initial topology, that we subsequently prune by removing edges longer than a threshold distance $d_{\min}$. 
We use Euclidean distances between nuclei centroids in the image space to quantify cellular distances.
Formally, for each node $v$, an edge $e_{vu}$ is built if
\begin{equation}
\begin{split}
    u \in \{ w \; | \; \dist(v, w) \leq d_k \wedge \dist(v, w) < d_{\min}, \; \forall w, v \in V_{\cg}, \; \\, d_k= k^\mathrm{th}\text{ smallest distance in } \dist(v, w)\}
\end{split}
\end{equation}
$\cg$ topology is represented by a binary adjacency matrix $E_{\cg} \in \mathbb{R}^{\;|V_{\cg}| \; \times \; |V_{\cg}|}$. Figure~\ref{fig:graph_generation} illustrates the $\cg$ representation for a sample $\troi$. Formally, a $\cg$ representation is formulated as $G_\cg :=$\{$V_{\cg}, E_{\cg}, H_{\cg}$\}.

\subsubsection{Tissue-graph representation}
\label{sec:tissue_graph_representation}
A tissue graph ($\tg$) depicts a high-level tissue microenvironment, where the nodes and edges of a $\tg$ denote tissue regions and their interactions, respectively. Similarly to a $\cg$, a $\tg$ is constructed by first identifying tissue regions (\eg epithelium, stroma, lumen, necrosis, and others in this paper), followed by feature representation of such tissue regions, and finally the $\tg$ topology construction. The steps are illustrated in Figure~\ref{fig:graph_generation}. 

Tissue regions are identified in a two-step process. First, we oversegment the tissue at a low magnification to detect non-overlapping homogeneous superpixels. Operating at low magnification avoids noisy pixels and provides computational efficiency. To this end, we employ the Simple Linear Iterative Clustering (SLIC) superpixel algorithm~\cite{achanta12}. Formally, SLIC follows an unsupervised approach by associating each pixel with a feature vector and merging the pixels using a localized version of k-means clustering. 
Next, we iteratively merge neighboring superpixels that have similar color attributes, \ie channel-wise mean and standard deviation, to create superpixels that capture meaningful tissue information. A sample tissue-region instance-map is presented in Figure~\ref{fig:graph_generation}.

To extract feature representations of tissue regions, we follow a two-step procedure: First, we extract $\cnn$-based features for oversegmented superpixels. Patches of size $h \times w$ centered around oversegmented superpixel centroids are processed by ResNet.
Second, morphological features of a tissue region are obtained by averaging the deep features of its constituting superpixels. Similar to $\cg$, we include spatial features as the normalized centroids of the tissue region. For a $\troi$ with a set of $V_{\tg}$ tissue regions, we denote the $\tg$ node-feature matrix as $H_{\tg} \in \mathbb{R}^{\; |V_{\tg}| \, \times \, d_{\tg}}$.

We assume adjacent tissue regions to biologically interact the most and thus be connected in the $\tg$ topology. To this end, we construct a Region Adjacency Graph~\cite{potjer96} where an edge is built between each adjacent tissue region. The $\tg$ topology is denoted by a binary adjacency matrix $E_{\tg} \in \mathbb{R}^{\;|V_{\tg}| \, \times \, |V_{\tg}|}$.
Formally, a $\tg$ representation is formulated as $G_\tg :=$ \{$V_{\tg}, E_{\tg}, H_{\tg}$\}. 

\begin{figure*}[!t]
\centering
\centerline{\includegraphics[width=0.98\textwidth]{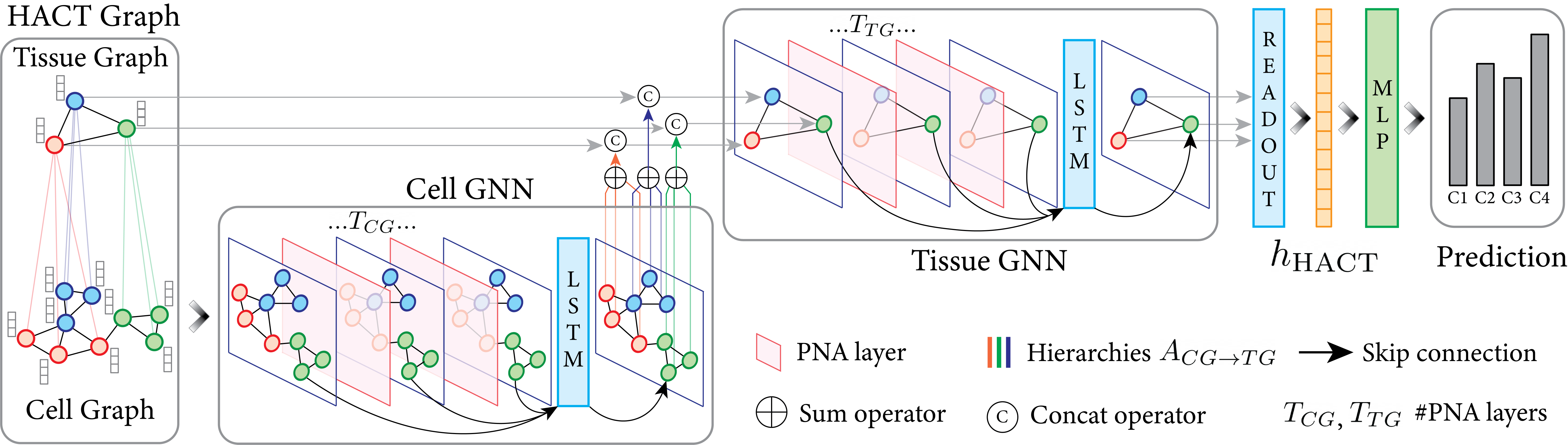}}
\caption{Overview of the proposed $\hact$-Net architecture. The network processes an input $\hact$ graph representation in a hierarchical manner, from finer cell-level to coarser tissue-region level, to obtain a contextualized graph embedding, which is subsequently used to classify the input graph. (Figure is best viewed in color)}
\label{fig:network}
\end{figure*}

\subsubsection{Hierarchical Cell-to-Tissue graph representation}
\label{sec:hierarchical_graph_representation}
Tissues in histopathology can be considered as hierarchical organizations of biological entities ranging from fine-level, \ie cells, to coarse-level, \ie tissue regions. There exist intra- and inter-level coupling based on topological distributions and interactions among the entities. 
Following this motivation, we propose $\hact$, a HierArchical Cell-to-Tissue ($\hact$) graph representation to jointly represent low-level $\cg$ and high-level $\tg$. Intra-level topology is already captured by $\cg$ and $\tg$ standalone. Inter-level topology is presented by a binary assignment (cell-to-tissue hierarchy) matrix $A_{\cg \rightarrow \tg} \in \mathbb{R}^{\;|V_{\cg}| \; \times \; |V_{\tg}|}$ that utilizes the relative spatial distributions of nuclei with respect to tissue regions. For the $i^\text{\,th}$ nucleus and $j^\text{\,th}$ tissue region, the corresponding assignment is given as
\begin{equation}
\begin{split}
   & A_{\cg \rightarrow \tg}[\,i, j\,] = 1, \; \text{if} \, i^\text{\,th} \, \text{nucleus centroid} \in j^\text{\,th} \, \text{tissue region} \\
   & A_{\cg \rightarrow \tg}[\,i, j\,] = 0, \; \text{otherwise}
\end{split}
\end{equation}

Cell-to-tissue hierarchies for a tissue region are presented in Figure~\ref{fig:graph_generation}.
Each nucleus is assigned to one and only one tissue region. If a segmented nucleus is at the border of multiple tissue regions, the nucleus is assigned to the tissue region that it has the maximum overlap with.
Formally for a given $\troi$, a $\hact$ representation is formulated as
$G_\hact :=$ \{$G_{\cg}, G_{\tg}, A_{\cg \rightarrow \tg}$\}.

\subsection{Graph learning}
\label{sec:graph_learning}
The $\hact$ graph representation for a $\troi$ is processed by a hierarchical $\gnn$ to map the $\troi$ composition to respective $\troi$ subtype. To this end, we propose a HierArchical Cell-to-Tissue Network ($\hact$-Net), a hierarchical $\gnn$ architecture illustrated in Figure~\ref{fig:network}.

\subsubsection{HACT-Net architecture \& learning}
\label{sec:hact_net}
$\hact$-Net takes a $\hact$ representation $G_\hact$ as input and outputs a graph-level representation $h_\hact \in \mathbb{R}^{d_\hact}$. Subsequently, a multi-layer perceptron ($\mlp$) classifies $h_\hact$ into a respective tissue class, \eg cancer subtype. 
Formally, $\hact$-Net consists of two $\gnn$s, namely Cell-$\gnn$ ($\cg$-$\gnn$) and Tissue-$\gnn$ ($\tg$-$\gnn$), to hierarchically process the $\hact$ graph from fine to coarse level.
In this work, we leverage the recent advances in $\gnn$s and model $\hact$-Net using Principal Neighbourhood Aggregation ($\pna$) layers~\cite{corso20}.

First, $\cg$-$\gnn$ takes $G_{\cg} :=$ $\{V_{\cg}, E_{\cg}, H_{\cg} \}$, and applies $T_{\cg}$ $\pna$ layers to build contextualized cell-node embeddings.
Inline with Equation~\eqref{eqn:gnn_preliminary}, we iteratively update each node embedding $h^{(t)}(v)$, $\forall v \in V_{\cg}$ as
\begin{equation}
\label{eqn:cg_node_update}
\begin{split}
& a_{\,\cg}^{\, (t \,+\,1)}(v) =  \mathlarger{\mathlarger{\mathlarger{\oplus}}}_{\,u \, \in \, \mathcal{N}_{\,\cg}(v)} \, M_{\,\cg}^{\,(t)}\, \Big(\, h_{\,\cg}^{\,(t)}(v), \,h_{\,\cg}^{\,(t)}(u)\, \Big) \\
& h_{\,\cg}^{\, (t \,+\,1)}(v) = U_{\,\cg}^{\,(t)}\,\Bigg(\,h_{\,\cg}^{\,(t)}(v), \, a_{\,\cg}^{\, (t \,+\,1)}(v) \,\Bigg)
\end{split}
\end{equation}
where $t=0,\dots,T_{\cg}$ is the iteration index, $\mathcal{N}_{\,\cg}(v)$ is the set of cell-nodes neighboring $v$, and the functions $U_{\,\cg}^{\,t}$ and $M_{\,\cg}^{\,t}$ are $\mlp$s. $\mathlarger{\mathlarger{\mathlarger{\oplus}}}$~denotes the combination of multiple degree-scalers and aggregators, \ie
\begin{equation}
\label{eqn:pna}
\begin{split}
& \mathlarger{\mathlarger{\mathlarger{\oplus}}} = \Big[I, \mathcal{S}(D, \alpha=1), \mathcal{S}(D, \alpha=-1)\Big] \, \mathlarger{\mathlarger{\mathlarger{\otimes}}} \, \Big[\,\mu, \sigma, \mathrm{max}, \mathrm{min}\,\Big] 
\, \\
& \mathcal{S}(D,\, \alpha) = \frac{\log \; (D + 1)^{\,\alpha}}{\delta}\, \\
& \delta = \frac{1}{|\text{train}|} \sum_{i \, \in \, \text{train}} \log \; (d_i + 1)
\end{split}
\end{equation}
where $I$ is the identity matrix, $\mathcal{S}$ is the degree-scaler matrix, $D$ is the degree matrix of cell-nodes, $[\mu, \sigma, \mathrm{max}, \mathrm{min}]$ is the list of aggregators that compute statistics on neighboring nodes, $\mathlarger{\mathlarger{\mathlarger{\otimes}}}$ is tensor product, $\delta$ is a normalization constant computed as the average log-scale cell-node degree from the training dataset, and $\alpha$ is a variable (that is negative for attenuation, positive for amplification, or zero for no scaling). The schematic diagram of a $\pna$ layer is shown in Figure~\ref{fig:gin_pna}. 
After $T_\cg$ $\pna$ layers, an $\lstm$-based jumping knowledge technique~\cite{xu18} is employed to adapt to different $\cg$ sub-graph structures, \ie
\begin{equation}
\label{eqn:cg_aggregator}
\begin{split}
& h^{\,(T_\cg+1)}_\cg(v) = \lstm\,\Big(\,\Big\{\,h_{\,\cg}^{\,(t)}(v) \;\Big|\;t = 1, \dots , T_{\cg} \, \Big\} \, \Big)
\end{split}
\end{equation}

The cell node embeddings $h^{\,T_\cg+1}_\cg(v) \; | \; v \in V_\cg$ and the assignment matrix $A_{\cg \rightarrow \tg}$ are used as additional hierarchical information to initialize the tissue-node features in the $\tg$, \ie 
\begin{equation}
\label{eqn:tg_node_update}
h_{\,\tg}^{\, (0)}(w) = \concat\,\Big(\, H_{\,\tg}(w), \sum_{v \, \in \, \mathcal{M}\,(w)} h_\cg^{\,(T_\cg+1)}(v)\,\Big) \\
\end{equation}
where $\concat$ denotes the concatenation operation and ${\mathcal{M}}\,(w) :=$ $\{\,v \in V_\cg \; | \; A_{\cg \rightarrow \tg}\,(v, w) = 1 \, \}$ is the set of nodes in $G_\cg$ mapping to a node $w \in V_{\tg}$.
Analogously to Equation~\eqref{eqn:cg_node_update}, we process $G_\tg$ using the $\tg$-$\gnn$, based on $\pna$ layers, to compute tissue region node embeddings 
$h^{(t)}_\tg(w), \; \forall w \in V_\tg$. At $t=T_\tg$, the embedding of each tissue component node $w$ encodes the cell and tissue information up to $T_\tg$-hops from $w$. 



Similar to $\cg$, the tissue-node embeddings in $\tg$ are processed with an $\lstm$-based jumping knowledge technique to aggregate the intermediate tissue-node representations. Finally, the graph-level embedding $h_\hact$ is extracted by summing all tissue-node representations.
An $\mlp$ layer followed by softmax operation provides a mapping from $h_\hact$ to the respective $\troi$ label. The model is trained end-to-end to minimize the cross-entropy loss between the softmax output and the ground-truth $\troi$ label.


Following~\cite{dwivedi20}, after each $\pna$ layer we include graph normalization (GraphNorm) followed by a batch normalization (BatchNorm). Graph normalization scales the node features by the number of nodes in the graph. Intuitively, it prevents the node representations from being at different scales, for graphs of different sizes. This normalization helps the network to learn discriminative topological patterns when the number of nodes vary significantly within a class.

\begin{figure*}[!t]
\centering
\centerline{\includegraphics[width=0.82\textwidth]{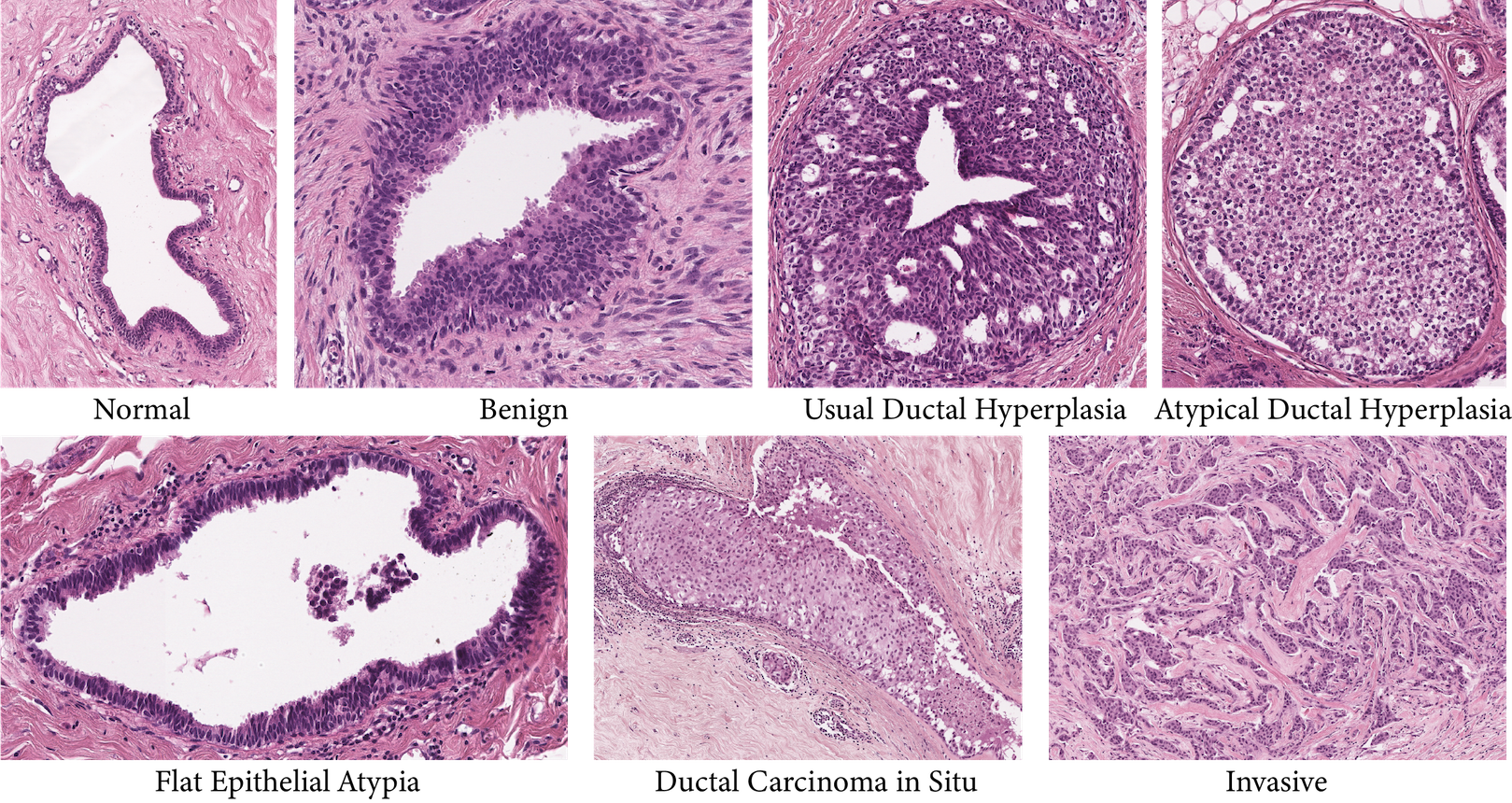}}
\caption{Samples of class-wise tumor regions-of-interest in BRACS dataset. (Figure is best viewed in color)}
\label{fig:samples}
\end{figure*}

\begin{figure*}[!t]
\centering
\centerline{\includegraphics[width=0.82\textwidth]{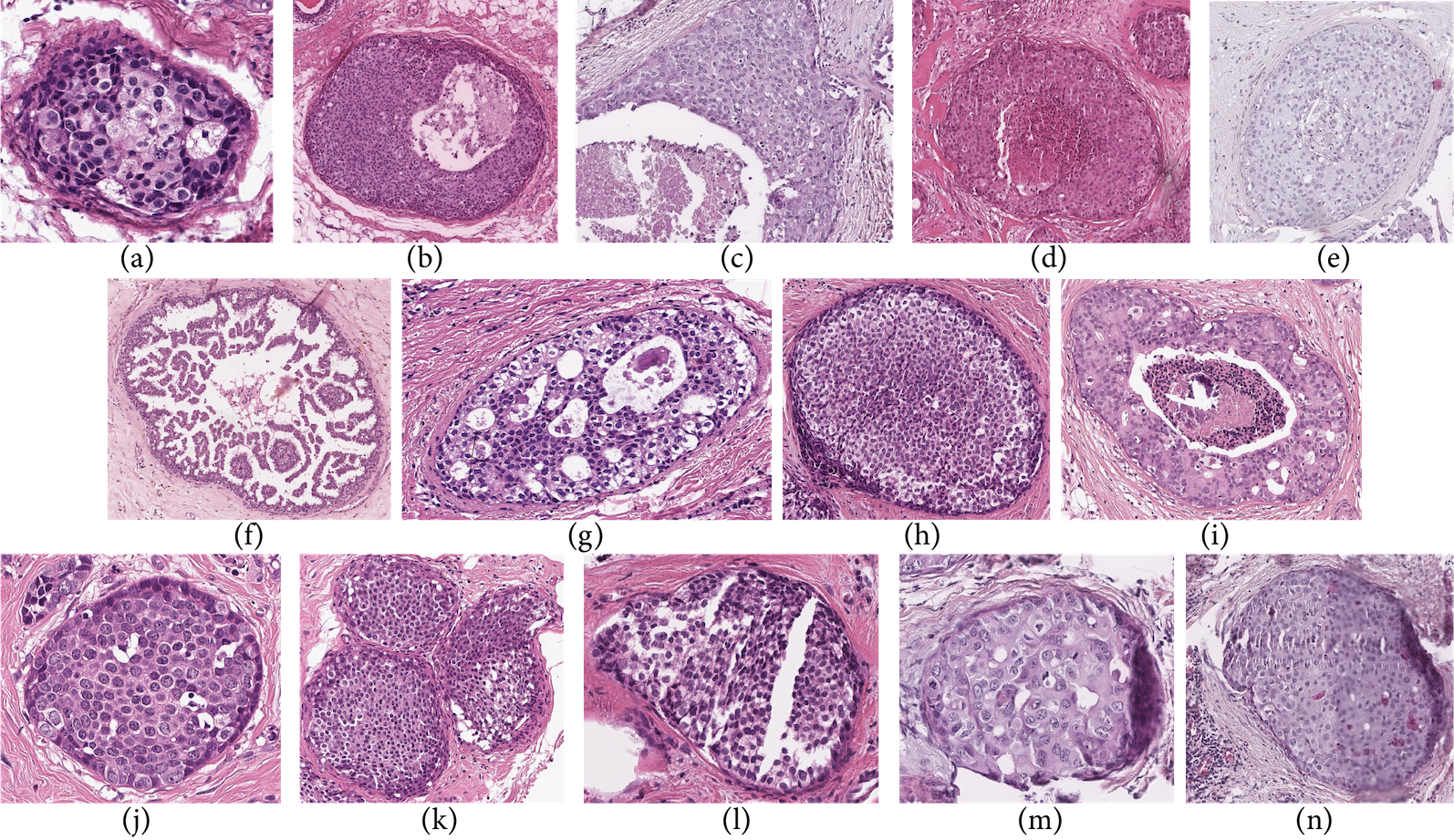}}
\caption{Overview of the variability in Ductal Carcinoma \textit{in situ} tumor regions-of-interest in BRACS dataset. (Figure is best viewed in color)}
\label{fig:variability}
\end{figure*}

\begin{table*}[t]
\caption{Key statistics of the BRACS dataset.}
\label{tab:dataset_statistics}
\centering
\scriptsize 
\begin{tabular}{c|lccccccc|c}
  \toprule
  & Metric & Normal & Benign & UDH & ADH & FEA & DCIS & Invasive & Total \\
  \midrule
  \parbox[t]{2mm}{\multirow{2}{*}[-1.4ex]{\rotatebox[origin=c]{90}{Image}}} & Number of images & 512 & 758 & 471 & 568 & 783 & 749 & 550 & 4391 \\ [0.1cm]
  
  & Number of pixels (in million) & 2.8$\pm$2.7 & 5.7$\pm$4.5 & 2.4$\pm$2.9 & 2.2$\pm$2.0 & 1.2$\pm$1.1 & 5.0$\pm$5.0 & 8.2$\pm$5.4 & 3.9$\pm$4.3 \\ [0.1cm]
  
  & Max/Min pixel ratio & 75.3 & 97.9 & 180.1 & 75.3 & 58.3 & 128.6 & 62.4 & 235.6 \\ [0.1cm]
  
  \midrule
  \parbox[t]{2mm}{\multirow{3}{*}[-0.8ex]{\rotatebox[origin=c]{90}{$\cg$}}} & Number of nodes & 994$\pm$732 & 1826$\pm$1547 & 903$\pm$910 & 863$\pm$730 & 470$\pm$352 & 1723$\pm$1598 & 3609$\pm$2393 & 1468$\pm$1642 \\ [0.1cm]
  
  & Number of edges & 3759$\pm$2643 & 6103$\pm$5420 & 3371$\pm$3675 & 3098$\pm$2781 & 1738$\pm$1395 & 5728$\pm$5811 & 12490$\pm$10011 & 5102$\pm$6089\\ [0.1cm]
  
  & Max/Min node ratio & 71.9 & 126.6 & 133.3 & 104.2 & 45.2 & 161.3 & 113.6 & 256.4 \\ [0.1cm]

  \midrule
  \parbox[t]{2mm}{\multirow{3}{*}[-0.8ex]{\rotatebox[origin=c]{90}{$\tg$}}} & Number of nodes & 107$\pm$106 & 217$\pm$233 & 88$\pm$93 & 100$\pm$91 & 45$\pm$32 & 225$\pm$217 & 423$\pm$317 & 172$\pm$217 \\ [0.1cm]
  
  & Number of edges & 509$\pm$545 & 1012$\pm$1236 & 393$\pm$450 & 480$\pm$474 & 194$\pm$159 & 1111$\pm$1123 & 2025$\pm$1741 & 815$\pm$1125 \\ [0.1cm]
  
  & Max/Min node ratio & 169.5 & 312.5 & 125.0 & 178.6 & 416.7 & 312.5 & 101.0 & 434.8 \\ [0.1cm]
  
  \midrule
  \parbox[t]{2mm}{\multirow{3}{*}[-0.3ex]{\rotatebox[origin=c]{90}{Image split}}} & Train & 342 & 586 & 303 & 405 & 599 & 562 & 366 & 3163 \\ [0.1cm]
  & Validation & 86 & 87 & 88 & 77 & 85 & 97 & 82 & 602 \\ [0.1cm]
  & Test & 84 & 85 & 80 & 86 & 99 & 90 & 102 & 626 \\ [0.1cm]
  
  \midrule
  \parbox[t]{2mm}{\multirow{3}{*}[-0.5ex]{\rotatebox[origin=c]{90}{WSI split}}} & Train & 67 & 86 & 59 & 38 & 37 & 33 & 41 & 198 \\ [0.1cm]
  & Validation & 28 & 24 & 24 & 28 & 17 & 21 & 19 & 68 \\ [0.1cm]
  & Test & 15 & 16 & 20 & 17 & 12 & 16 & 16 & 59 \\ [0.1cm]

  \bottomrule
\end{tabular}
\end{table*}

\section{Datasets}
\label{sec:dataset}

\textbf{BRACS dataset:}
As part of this work, we introduce a new dataset termed as the BReAst Cancer Subtyping (\textbf{BRACS}) dataset. BRACS contains 4391 $\troi$s acquired from 325 H\&E stained breast carcinoma WSIs. The WSIs were selected from the archives of the Department of Pathology at National Cancer Institute- IRCCS-Fondazione Pascale, Naples, Italy, and are scanned with an Aperio AT2 scanner at 0.25 $\mu$m/pixel resolution.
The $\troi$s were selected and annotated using QuPath software~\cite{bankhead17} as being Normal, Benign, Usual ductal hyperplasia (UDH), Atypical Ductal Hyperplasia (ADH), Flat Epithelial Atypia (FEA), Ductal Carcinoma In Situ (DCIS), and Invasive. 
Figure~\ref{fig:samples} presents sample $\troi$s from all the cancer subtypes in BRACS. 

Each $\troi$ was first annotated independently by three pathologists, and then any $\troi$s with any conflict in annotation were further discussed and annotated based on consensus of the three pathologists. Note that the pathologists utilize the entire WSI-level context during the $\troi$ annotation.
Figure.~\ref{fig:variability} demonstrates a few DCIS samples in BRACS to depict the appearance variability per category. Figure~\ref{fig:variability}\textcolor{blue}{(a,b,c)} display the variability in the $\troi$ sizes. Figure~\ref{fig:variability}\textcolor{blue}{(d,e)} display the variability in staining appearance. Figure~\ref{fig:variability}\textcolor{blue}{(f,g,h,i)} display different patterns of low-grade (Papillary), moderate-grade (Cribriform) and high-grade (Solid and Comedo) DCIS in the dataset. Figure~\ref{fig:variability}\textcolor{blue}{(j,k)} present DCIS $\troi$s with single and multiple glandular regions. Figure~\ref{fig:variability}\textcolor{blue}{(l,m,n)} presents some notable artifacts in tissue and slide preparation, such as tissue-folding, tears, ink stains, and blur. Similar $\troi$ variability is typical in practice and it exists in our dataset for other cancer subtypes as well.
Including such natural variability in the dataset (instead of ``cleaning'' them, \ie filtering them out) ensures for us a realistic and representative evaluation set, with results readily applicable in the field.

Table~\ref{tab:dataset_statistics} presents category-wise statistics of the $\troi$s in BRACS. The statistics demonstrate a high variation in $\troi$ dimensions.
Additionally, we also present the statistics for the $\cg$ and $\tg$ representations constructed by our framework as described earlier above. These indicate a large variation in the size of the constructed graph representations.
For evaluations on BRACS, we partition the $\troi$s into train, validation, and test sets at the WSI-level, such that two $\troi$s from the same WSI do not fall in different sets. The WSI-level splitting was performed randomly, ensuring a comparable number of $\troi$s per cancer subtype. Such partitioning aimed for a fair evaluation of the compared methods.

\textbf{BACH dataset:}
We evaluated the proposed methodology also on the publicly available microscopy image dataset called the Grand Challenge on BreAst Cancer Histology images \textbf{BACH}~\cite{aresta19}. This dataset consists of 400 training and 100 test images from four breast cancer subtypes, \ie Normal, Benign, DCIS, and Invasive. All images are acquired using a Leica DM 2000 LED microscope and a Leica ICC50 HD camera. These images are in RGB TIFF format and have a fixed size of 2048$\times$1536 pixels and a pixel scale of 0.42$\times$0.42\,$\mu$m. 
Notably, our BRACS dataset has three major advantages over the BACH dataset:
\vspace{-\topsep}
\begin{itemize}[noitemsep]
    \item Number of images: The train and test sets of BRACS are nearly 10 times and 6 times the size of the train and test sets of BACH, respectively. The large test set ensures a robust evaluation of the methods.
    \item Diverse subtypes: BRACS includes diagnostically complex pre-cancerous atypical (ADH and FEA) categories, which represent a major diagnostic dilemma typical in practice, due to their high risk of progressing to cancer. The seven cancer subtypes in BRACS represent a broad spectrum of breast cancer in histopathology.
    \item Large variability: The aforementioned high variability in the BRACS dataset in terms of $\troi$ appearances and dimensions is clinically more representative and corresponds to a more realistic scenario of breast cancer subtyping.
\end{itemize}

\section{Results}
\label{sec:results}

In this section, we comparatively evaluate the proposed methods for breast cancer subtyping. 
First, we introduce state-of-the-art $\cnn$ and $\gnn$ baselines and their implementation strategies.
Second, we conduct ablation studies on BRACS dataset to examine the impact of various components of our proposed framework. 
Third, we evaluate the classification performance of our method, comparatively with the given baselines, on BRACS and BACH datasets for different classification settings.
Finally, we include a comparison with gold-standard, by analyzing the performance of $\hact$-Net against three independent expert-pathologists.

\subsection{$\cnn$ and $\gnn$ baselines for comparative evaluation}
\label{sec:baselines}

\textbf{$\bullet$ Single-scale CNN} 
processes $\troi$s at a single magnification. A $\cnn$ is trained to predict patch-wise cancer subtypes and we aggregate the patch-wise predictions to produce a $\troi$-level prediction.
We experiment with images at three magnifications, namely 10$\times$, 20$\times$, and 40$\times$, denoted herein as $\cnn$(10$\times$), $\cnn$(20$\times$), and $\cnn$(40$\times$). Same network architecture and training strategy is employed for all scales.
For each scale, we extract patches of size 128$\times$128 pixels with a stride of 64 pixels.
The $\cnn$ follows the single-scale training procedure by~\cite{sirinukunwattana18} and patch-wise predictions are aggregated using the Agg-Penultimate strategy proposed by~\cite{mercan19}.
We use transfer learning with a ResNet-50 architecture, pre-trained on ImageNet dataset, as our $\cnn$ backbone. Following feature extraction by ResNet-50, we employ a two-layer $\mlp$ of 128 channels to classify the patches. To improve classification performance, we used fine-tuning from the original ResNet-50 parameters. Adam optimizer~\cite{kingma15} with a learning rate of $10^{-3}$, a batch size of 16, and a dropout ratio of 0.2 were used to optimize the categorical cross-entropy objective.

\textbf{$\bullet$ Multi-scale CNN}
processes the $\troi$s at multiple scales. We extract concentric patches of size 128$\times$128 pixels from multiple magnifications and follow the ``Late fusion with single-stream + $\lstm$" training procedure from~\cite{sirinukunwattana18}. This multi-scale approach uses concentric patches to acquire context information from multiple magnifications for improving the patch classification. 
We operate at two settings, \ie (10$\times$+20$\times$) and (10$\times$+20$\times$+40$\times$), and denoted by prepending Multi-scale $\cnn$ in front of each. 
The patch-wise predictions are aggregated using the Agg-Penultimate strategy by~\cite{mercan19}.
On the concatenated feature representation from the $\lstm$, we employ a two-layer $\mlp$ of 128 channels to classify the patches. The training strategy and hyperparameters are the same as Single-scale CNN.

\textbf{$\bullet$ CGC-Net}
is the Cell Graph Convolutional Network (CGC-Net) proposed by~\cite{zhou19}, and it is the state-of-the-art in classifying $\cg$ representations for $\troi$s. We construct the $\cg$ topology for a $\troi$ using thresholded kNN strategy presented in Section~\ref{sec:cell_graph_representation}. We initialize the $\cg$ nodes with hand-crafted features, employ the Adaptive GraphSage-based CGC-Net architecture, and follow the training strategy proposed by~\cite{zhou19}.

\textbf{$\bullet$ Patch-GNN}
implements the methodology proposed by~\cite{aygunes20}, which is the state-of-the-art $\gnn$ method for classifying patch-graph representations of $\troi$s. This methodology incorporates local inter-patch context through a $\gnn$ to construct a graph-level feature representation, which is then processed by an $\mlp$ to classify the $\troi$s. 
We experiment with Patch-$\gnn$ at three scales, \ie 10$\times$, 20$\times$, and 40$\times$, denoted herein as Patch$-\gnn$(10$\times$), Patch$-\gnn$(20$\times$), and Patch$-\gnn$(40$\times$). 
At each magnification, we extract patches of size 128$\times$128 to construct a $\troi$-specific patch-graph. We employ the network architecture and training strategy proposed by~\cite{aygunes20}.

\textbf{$\bullet$ CG-GNN}
is provided as a standalone cell-graph based learning baseline, to compare with our proposed hierarchical learning. 
$\cg$-$\gnn$ architecture utilizes $\pna$ layers, an $\lstm$-based jumping knowledge, sum readout, and a two-layer $\mlp$ classifier.
We follow the same $\cg$ representation strategy as described above in Section~\ref{sec:cell_graph_representation}.

\textbf{$\bullet$ TG-GNN}
is provided as a standalone tissue graph-based learning baseline, to compare with our proposed hierarchical learning. 
$\tg$-$\gnn$ employs the same architecture as the $\cg$-$\gnn$, with the node features directly initialized by $H_{\,\tg}$ instead of Equation~\eqref{eqn:tg_node_update}.

\textbf{$\bullet$ CONCAT-GNN}
is provided to evaluate the impact of hierarchical graph representation and learning. $\concat$-$\gnn$ utilizes standalone $\cg$ and $\tg$ representations, respectively, as input to standalone $\cg$-$\gnn$ and $\tg$-$\gnn$ to produce $h_{\cg}$ and $h_{\tg}$ graph-level embeddings. 
The $\troi$ level embedding is constructed by concatenating the graph-level embeddings, \ie $h_{\,\concat}$ = $\concat(\,h_{\cg}, \, h_{\,\tg})$. Finally, a two-layer $\mlp$ classifies $h_{\,\concat}$ into a cancer subtype.

\subsection{Implementation}
\label{sec:implementation}
\textbf{Graph representations:}
$\cg$ representations (Section~\ref{sec:cell_graph_representation}) use, i)~patches of size 72$\times$72, and ii)~a $\cnn$ of ResNet-34 or ResNet-50 to initialize the node features.
$\tg$ representations (Section~\ref{sec:tissue_graph_representation}) use, i)~patches of size 144$\times$144, and ii)~a $\cnn$ of ResNet-34 or ResNet-50 to initialize the node features.

\textbf{Graph architecture and learning:}
$\cg$-$\gnn$, $\tg$-$\gnn$, $\concat$-$\gnn$, and $\hact$-Net all share the same options and hyperparameters below
\vspace{-\topsep}
\begin{itemize}[noitemsep]
\item \# $\pna$ layers in $\gnn$: [3, 4, 5]
\item \# $\mlp$ layers in a $\pna$ layer: 2
\item \# channels in a $\pna$-layer $\mlp$: 64
\item Graph-level embedding dimension: 128
\item \# $\mlp$ layers in output classifier: 2
\item \# channels in output $\mlp$ classifier: 128
\item Training parameters: Adam optimizer~\cite{kingma15} with a learning rate of $10^{-3}$, batch size of 16, and a categorical cross-entropy objective.
\end{itemize}

\textbf{Evaluation metrics:}
Considering the imbalanced number of $\troi$s per class in train, validation, and test sets (see Table~\ref{tab:dataset_statistics}), we evaluate the classification performance using the weighted F1-score, an average weighted by the number of true instances for each class. 
During the training of each method, the model with the best weighted F1-scores in the validation set is selected as the final trained model of that method. 
To present any sensitivity to initialization, we report the mean and standard deviation of each model on the test set, by training them three times using random weight initialization. Further, we present precision, recall, and confusion matrices to indicate the distribution of class predictions.

\textbf{Computational resources:}
All the experiments were conducted using PyTorch~\cite{paszke19} and Deep Graph Library (DGL)~\cite{wang19}, on NVIDIA Tesla P100 GPUs and POWER8 processors.

\subsection{Ablation studies}
\label{sec:ablation_studies}
We conduct ablation studies to evaluate the impact of three major components of our proposed methodology on $\troi$ classification performance, namely i)~node feature initialization, ii)~$\gnn$ layer type, and iii)~jumping knowledge technique. Each component is analyzed individually, while fixing the others. Ablation studies were performed on the BRACS dataset for classifying the $\troi$s into 7-classes.

\subsubsection{Impact of node feature initialization}
\label{sec:node_features}
The performance of $\gnn$s eminently rely on the initial node features~\cite{kipf17}. In the context of entity-based tissue analysis, we analyze the impact of initial morphological features of the nodes. To this end, we experiment with the following three morphological feature initialization schemes:

\textbf{$\bullet$ No morphological features:} In this setting, the nodes of an entity-graph representation are initialized with only the spatial features. Experiments with this setting demonstrate the impact of standalone graph topology on the $\gnn$ performance.

\begin{table}[t]
\caption{Ablation: Impact of node features. Mean and standard deviation of 7-class weighted F1-scores. Results expressed in $\%$.}
\label{tab:node_features}
\centering
\scriptsize 
\begin{tabular}{lc}
  \toprule
   & Weighed F1\\
  \midrule
  CG-$\gnn$: No morphological features & 45.24$\pm$1.51 \\ [0.1cm]
  CG-$\gnn$: Hand-crafted morphological features & 48.34$\pm$5.22 \\ [0.1cm]
  CG-$\gnn$: $\cnn$ morphological features & \textbf{55.94$\pm$1.01} \\ [0.1cm]
  \midrule
  TG-$\gnn$: No morphological features & 36.81$\pm$0.71 \\ [0.1cm]
  TG-$\gnn$: Hand-crafted morphological features & 51.62$\pm$2.11 \\ [0.1cm]
  TG-$\gnn$: $\cnn$ morphological features & \textbf{56.62$\pm$1.35} \\ [0.1cm]
  \midrule
  CONCAT-$\gnn$: No morphological features & 47.62$\pm$1.56 \\ [0.1cm]
  CONCAT-$\gnn$: Hand-crafted morphological features & 51.55$\pm$1.32 \\ [0.1cm]
  CONCAT-$\gnn$: $\cnn$ morphological features & \textbf{57.01$\pm$2.27} \\ [0.1cm]
  \midrule
  $\hact$-Net: No morphological features & 48.70$\pm$0.16 \\ [0.1cm]
  $\hact$-Net: Hand-crafted morphological features & 52.46$\pm$0.19 \\ [0.1cm]
  $\hact$-Net: $\cnn$ morphological features & \textbf{61.53$\pm$0.87} \\
  \bottomrule
\end{tabular}
\end{table}

\textbf{$\bullet$ Hand-crafted morphological features:} The entity-graph nodes are initialized with hand-crafted morphological features as suggested by~\cite{zhou19}, \ie i)~\emph{texture features}: difference of average foreground to background; standard deviation, skewness, and mean entropy of intensities; dissimilarity, homogeneity, energy, and angular second moment from Gray-Level Co-occurrence Matrix; and ii)~\emph{shape features}: eccentricity, area, maximum and minimum axis lengths, perimeter, solidity, and orientation. Note that, the hand-crafted features for $\cg$ and $\tg$ are computed, respectively, from the segmented instances of nuclei and tissue regions. 

\textbf{$\bullet$ CNN morphological features:} The morphological features of the graph nodes are initialized with $\cnn$ features (ResNet-34 pre-trained on ImageNet) extracted from patches around the centroids of the nuclei and tissue regions.

Experimental results in Table~\ref{tab:node_features} indicate that the standalone $\cg$ topology is more discriminative for cancer subtyping than $\tg$ topology. The combination of $\cg$ and $\tg$ topologies further improves discriminative ability. The best performance achieved with the $\hact$ topology confirms the strength of hierarchical representations.
Further, including morphological features significantly improves the discriminative ability. The superiority of graphs with $\cnn$-based morphological features indicate the richness of morphological information acquired by $\cnn$s, compared to hand-crafted measures.

\begin{table}[t]
\caption{Ablation: Impact of $\gnn$ layer. Mean and standard deviation of 7-class weighted F1-scores. Results expressed in $\%$.}
\label{tab:gnn_layer}
\centering
\scriptsize 
\begin{tabular}{lc}
  \toprule
   & Weighed F1\\
  \midrule
  CG-$\gnn$: GIN & 55.70$\pm$0.51 \\ [0.1cm]
  CG-$\gnn$: PNA & \textbf{55.94$\pm$1.01} \\ [0.1cm]
  \midrule
  TG-$\gnn$: GIN & 55.33$\pm$1.36 \\ [0.1cm]
  TG-$\gnn$: PNA & \textbf{56.62$\pm$1.35} \\ [0.1cm]
  \midrule
  CONCAT-$\gnn$: GIN & 56.20$\pm$2.12 \\ [0.1cm]
  CONCAT-$\gnn$: PNA & \textbf{57.01$\pm$2.27} \\ [0.1cm]
  \midrule
  $\hact$-Net: GIN & 59.73$\pm$1.20 \\ [0.1cm]
  $\hact$-Net: PNA & \textbf{61.53$\pm$0.87} \\ 
  \bottomrule
\end{tabular}
\end{table}

\begin{table}[t]
\caption{Ablation: Impact of $\gnn$ jumping knowledge technique. Mean and standard deviation of 7-class weighted F1-scores. Results expressed in $\%$.}
\label{tab:aggregators}
\centering
\scriptsize 
\begin{tabular}{lc}
  \toprule
   & Weighed F1\\
  \midrule
  CG-$\gnn$: No aggregator & 55.53$\pm$0.75 \\ [0.1cm]
  CG-$\gnn$: Concatenation & 55.82$\pm$0.97 \\ [0.1cm]
  CG-$\gnn$: LSTM & \textbf{55.94$\pm$1.01} \\ [0.1cm]
  \midrule
  TG-$\gnn$: No aggregator & 55.30$\pm$0.81 \\ [0.1cm]
  TG-$\gnn$: Concatenation & 56.07$\pm$0.80 \\ [0.1cm]
  TG-$\gnn$: LSTM & \textbf{56.62$\pm$1.35} \\ [0.1cm]
  \midrule
  CONCAT-$\gnn$: No aggregator & \textbf{57.67$\pm$4.66} \\ [0.1cm]
  CONCAT-$\gnn$: Concatenation & 56.28$\pm$2.75 \\ [0.1cm]
  CONCAT-$\gnn$: LSTM & 57.01$\pm$2.27 \\ [0.1cm]
  \midrule
  $\hact$-Net: No aggregator & 49.16$\pm$1.15 \\ [0.1cm]
  $\hact$-Net: Concatenation & 59.78$\pm$1.59 \\ [0.1cm]
  $\hact$-Net: LSTM & \textbf{61.53$\pm$0.87} \\ 
  \bottomrule
\end{tabular}
\end{table}

\subsubsection{Impact of $\gnn$ layer type}
\label{sec:gnn_layer}
We investigate the impact of two state-of-the-art $\gnn$ layers, \ie $\gin$ and $\pna$ (Figure~\ref{fig:gin_pna}), on the classification performance. The experiments use $\cnn$-based node feature initialization and $\lstm$-based jumping knowledge. Results in Table~\ref{tab:gnn_layer} demonstrate that $\gnn$s with $\pna$ layers outperform $\gnn$s with $\gin$ layers, for all the four $\gnn$ constructions. 

\begin{table*}[t]
\caption{Mean and standard deviation of per-class F1-scores and weighted F1-scores for 7-class classification setting. Results are expressed in $\%$. The best result is in \textbf{bold} and the second best is \underline{underlined}.}
\label{tab:7_class_results}
\centering
\scriptsize 
\begin{tabular}{c|lccccccc|c}
  \toprule
  & Method & Normal & Benign & UDH & ADH & FEA & DCIS & Invasive & Weighted F1 \\
  \midrule
  \parbox[t]{2mm}{\multirow{5}{*}{\rotatebox[origin=c]{90}{$\cnn$}}} & $\cnn(10 \times)$ & 48.67$\pm$1.71 & 44.33$\pm$1.89 & \underline{45.00$\pm$4.97} & 24.00$\pm$2.83 & 47.00$\pm$4.32 & 53.33$\pm$2.62 & \underline{86.67$\pm$2.64} & 50.85$\pm$2.64\\ [0.1cm]
  
  & $\cnn(20 \times)$ & 42.00$\pm$2.16 & 42.33$\pm$3.09 & 39.33$\pm$2.05 & 22.67$\pm$2.49 & 47.67$\pm$1.25 & 50.33$\pm$3.09 & 77.00$\pm$1.41 & 46.85$\pm$2.19 \\ [0.1cm]

  & $\cnn(40 \times)$ & 32.33$\pm$4.64 & 39.00$\pm$0.82 & 23.67$\pm$1.70 & 18.00$\pm$0.82 & 37.67$\pm$2.87 & 47.33$\pm$2.05 & 70.67$\pm$0.47 & 39.41$\pm$1.89 \\ [0.1cm]

  & Multi-scale $\cnn (10\times+20\times)$ & 48.33$\pm$2.05 & 45.67$\pm$0.47 & 41.67$\pm$4.99 & 32.33$\pm$0.94 & 46.33$\pm$1.41 & 59.33$\pm$2.05 & 85.67$\pm$1.89 & 52.27$\pm$1.93 \\ [0.1cm]

  & Multi-scale $\cnn (10\times+20\times+40\times)$ & 50.33$\pm$0.94 & 44.33$\pm$1.25 & 41.33$\pm$2.49 & 31.67$\pm$3.30 & 51.67$\pm$3.09 & 57.33$\pm$0.94 & 86.00$\pm$1.41 & 52.83$\pm$1.92 \\ [0.1cm]

  \midrule
  \parbox[t]{2mm}{\multirow{4}{*}{\rotatebox[origin=c]{90}{$\gnn$}}} & CGG-Net & 30.83$\pm$5.33 & 31.63$\pm$4.66 & 17.33$\pm$3.38 & 24.50$\pm$5.24 & 58.97$\pm$3.56 & 49.36$\pm$3.41 & 75.30$\pm$3.20 & 43.63$\pm$0.51 \\ [0.1cm]
  
  & Patch-$\gnn (10\times)$ & 52.53$\pm$3.27 & 47.57$\pm$2.25 & 23.67$\pm$4.65 & 30.66$\pm$1.79 & 60.73$\pm$5.35 & 58.76$\pm$1.15 & 81.63$\pm$2.17 & 52.10$\pm$0.61 \\ [0.1cm]
  
  & Patch-$\gnn (20\times)$ & 43.86$\pm$4.23 & 43.37$\pm$3.21 & 19.47$\pm$2.31 & 25.73$\pm$2.87 & 55.57$\pm$2.08 & 52.86$\pm$1.85 & 79.20$\pm$1.04 & 47.10$\pm$0.70 \\ [0.1cm]
  
  & Patch-$\gnn (40\times)$ & 41.70$\pm$3.06 & 32.93$\pm$1.04 & 25.07$\pm$3.74 & 25.63$\pm$2.01 & 49.47$\pm$3.46 & 48.60$\pm$4.23 & 71.57$\pm$5.15 & 43.23$\pm$0.57 \\ [0.1cm]
  
  \midrule
  \parbox[t]{2mm}{\multirow{4}{*}{\rotatebox[origin=c]{90}{Ours}}} & CG-$\gnn$ & 58.77$\pm$6.82 & 40.87$\pm$3.05 & \textbf{46.82$\pm$1.95} & \underline{39.99$\pm$3.56} & 63.75$\pm$10.48 & 53.81$\pm$3.89 & 81.06$\pm$3.33 & 55.94$\pm$1.01 \\ [0.1cm]
  
  & TG-$\gnn$ & \textbf{63.59$\pm$4.88} & \textbf{47.73$\pm$2.87} & 39.41$\pm$4.70 & 28.51$\pm$4.29 & \underline{72.15$\pm$1.35} & 54.57$\pm$2.23 & 82.21$\pm$3.99 & 56.62$\pm$1.35 \\ [0.1cm]
 
  & CONCAT-$\gnn$ & 60.97$\pm$4.54 & 43.06$\pm$2.26 & 41.96$\pm$4.67 & 26.10$\pm$3.73 & 71.29$\pm$2.09 & \underline{60.83$\pm$3.71} & 85.42$\pm$2.70 & \underline{57.01$\pm$2.27} \\ [0.1cm]
 
  & $\hact$-Net (Proposed) & \underline{61.56$\pm$2.15} & \underline{47.49$\pm$2.94} & 43.60$\pm$1.86 & \textbf{40.42$\pm$2.55} & \textbf{74.22$\pm$1.41} & \textbf{66.44$\pm$2.57} & \textbf{88.40$\pm$0.19} & \textbf{61.53$\pm$0.87} \\

  \bottomrule
\end{tabular}
\end{table*}

\begin{figure*}[t]
\centering
\begin{subfigure}[t]{.49\textwidth}
\centering
\includegraphics[width=\linewidth]{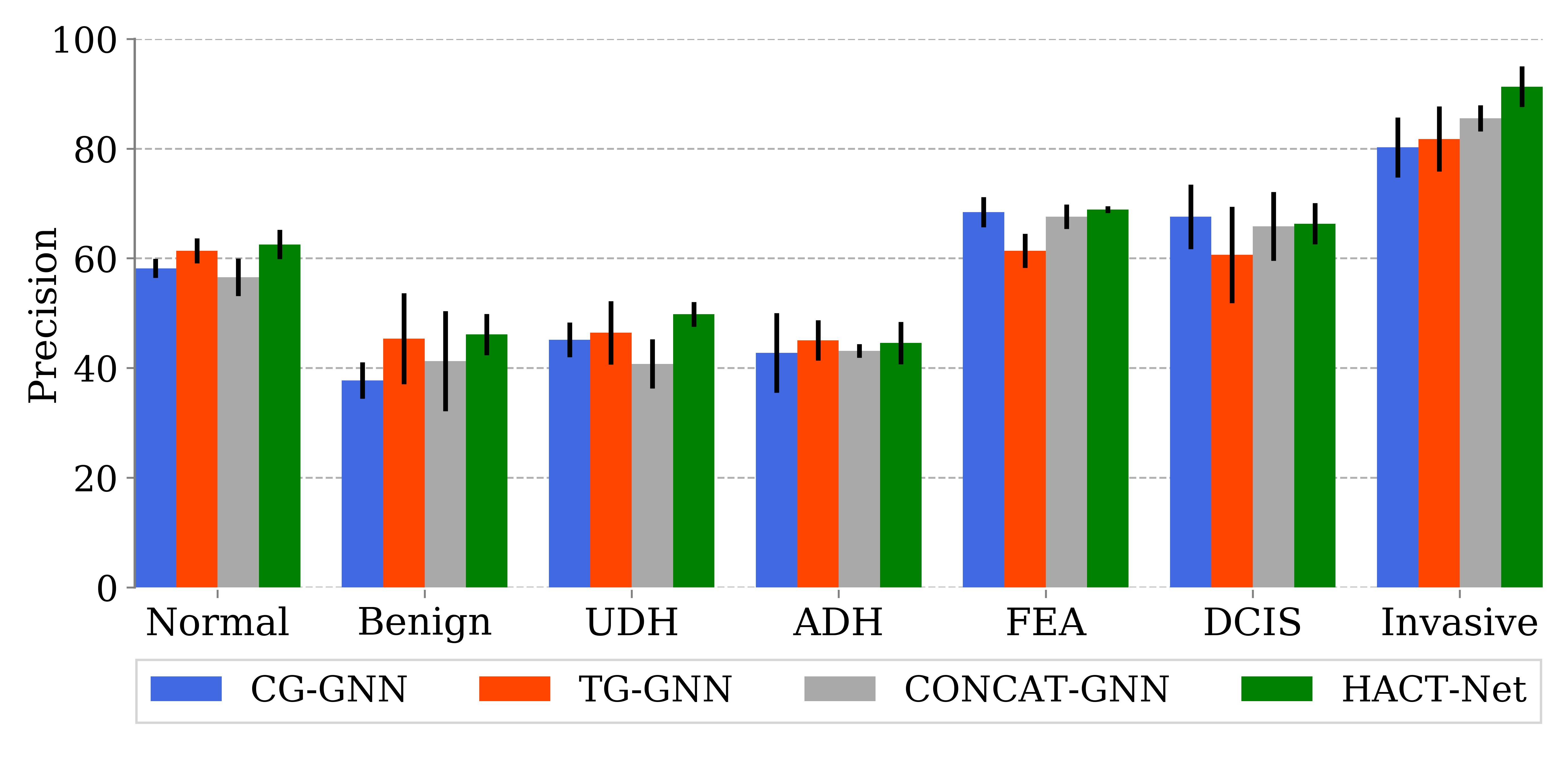}
\caption{}
\end{subfigure}
\begin{subfigure}[t]{.49\textwidth} 
\centering
\includegraphics[width=\linewidth]{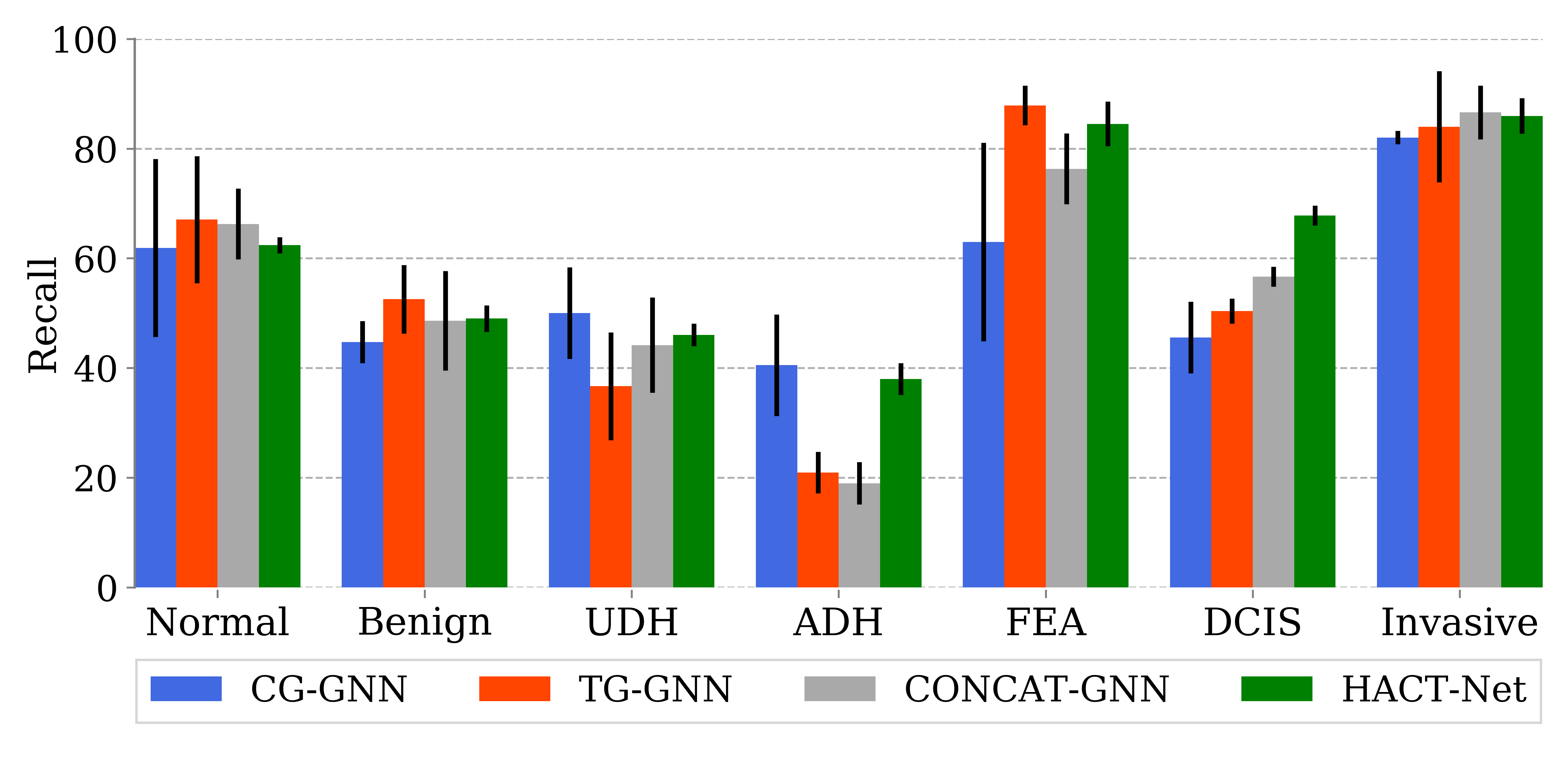}
\caption{}
\end{subfigure}
\caption{Mean and standard deviation of per-class precision and recall for 7-class classification setting. (Figure is best viewed in color)}
\label{fig:precision_recall}
\end{figure*}

\subsubsection{Impact of jumping knowledge technique}
\label{sec:gnn_aggregator}
To investigate the impact of the jumping knowledge technique, we experiment with three settings: no jumping knowledge, $\concat$-based, and $\lstm$-based. $\lstm$-based technique follows Equation~\eqref{eqn:cg_aggregator}. Based on this, $\concat$-based technique replaces the $\lstm$ operation with the concatenation operation.
The experiments use $\cnn$-based node feature initialization and $\pna$ layers. Results in Table~\ref{tab:aggregators} demonstrate a generally positive impact of the jumping knowledge technique. Compared to $\concat$, the $\lstm$-based technique learns better dependencies between $\gnn$ layers, thus generates better graph embeddings.

\subsubsection{Ablation summary}
\label{sec:ablation_summary}
The ablation experiments conclude the following choice of components for designing our methodology, i)~$\cnn$-based initialization of node-level morphological features, ii)~use of $\pna$ layers, and iii)~an $\lstm$-based jumping knowledge technique.

\subsection{Classification results on BRACS dataset}
\label{sec:bracs_results}
We evaluate our proposed methods, comparatively with $\cnn$ and $\gnn$ baselines. To analyze the performance for different clinical applications and histopathological needs, we evaluate and report the results separately in the following three settings:

\textbf{$\bullet$ Setting 1: 7-class classification:}
Here, we classify the $\troi$s into 7-classes, \ie Normal, Benign, UDH, ADH, FEA, DCIS, and Invasive, for the differentiation of a large spectrum of breast cancer subtypes. Table~\ref{tab:7_class_results} tabulates the classification performance of the compared methods. 

Among single-scale $\cnn$s, $\cnn$(10$\times$) is the best performing one, indicating the importance of global context information for $\troi$ classification. Multi-scale $\cnn$s using both global and local context information outperform single scale $\cnn$s. Such benefit from context is significant for ADH, FEA, and DCIS categories, which all require both local and global context information for the diagnosis. Multi-scale $\cnn$s also outperform CGC-Net and Patch-$\gnn$ baselines. Interestingly, at each magnification, Patch-$\gnn$ outperforms single-scale $\cnn$, which affirms the importance of relational and topological information incorporated in the form of patches.

\begin{table*}[t]
\caption{Mean and standard deviation of per-class F1-scores and weighted F1-scores for 4-class classification setting. Results are expressed in $\%$. The best result is in \textbf{bold} and the second best is \underline{underlined}.}
\label{tab:4_class_results}
\centering
\scriptsize 
\begin{tabular}{c|lcccc|c}
  \toprule
  & Method & Normal & Non-cancerous & Precancerous & Cancerous & Weighted F1 \\
  \midrule
  \parbox[t]{2mm}{\multirow{5}{*}{\rotatebox[origin=c]{90}{$\cnn$}}} & $\cnn(10 \times)$ & 54.33$\pm$3.68 & 56.00$\pm$0.82 & 56.33$\pm$1.25 & 83.67$\pm$0.94 & 64.36$\pm$1.37 \\ [0.1cm]

  & $\cnn(20 \times)$ & 45.33$\pm$4.64 & 55.33$\pm$0.47 & 52.33$\pm$1.89 & 81.67$\pm$2.05  & 61.18$\pm$1.93 \\ [0.1cm]

  & $\cnn(40 \times)$ & 42.00$\pm$4.89 & 51.00$\pm$0.82 & 47.67$\pm$4.11 & 77.67$\pm$2.05  & 56.99$\pm$2.72 \\ [0.1cm]

  & Multi-scale $\cnn (10\times + 20\times)$ & 51.67$\pm$5.79 & 55.33$\pm$1.25 & 52.67$\pm$2.87 & 80.67$\pm$1.89 & 61.82$\pm$2.53  \\ [0.1cm]

  & Multi-scale $\cnn (10\times + 20\times + 40\times)$ & 51.33$\pm$3.27 & 56.33$\pm$2.05 & 57.00$\pm$1.64 & 81.33$\pm$3.68 & 63.52$\pm$2.59  \\ [0.1cm]

  \midrule
  \parbox[t]{2mm}{\multirow{4}{*}{\rotatebox[origin=c]{90}{$\gnn$}}} & CGG-Net & 34.53$\pm$2.93 & 47.23$\pm$3.72 & 62.90$\pm$2.81 & 82.20$\pm$1.04 & 59.87$\pm$2.30  \\ [0.1cm]
  
  & Patch-$\gnn (10\times)$ & 53.13$\pm$4.40 & 46.23$\pm$2.45 & 63.96$\pm$3.82 & 77.43$\pm$3.22 & 61.93$\pm$2.51  \\ [0.1cm]
  
  & Patch-$\gnn (20\times)$ & 53.46$\pm$1.81 & 47.16$\pm$2.81 & 63.20$\pm$3.78 & 74.90$\pm$3.36 & 61.26$\pm$2.90  \\ [0.1cm]
  
  & Patch-$\gnn (40\times)$ & 40.90$\pm$2.75 & 38.67$\pm$2.76 & 56.77$\pm$3.91 & 72.20$\pm$2.61 & 54.60$\pm$1.90  \\ [0.1cm]
  
  \midrule
  \parbox[t]{2mm}{\multirow{4}{*}{\rotatebox[origin=c]{90}{Ours}}} & CG-$\gnn$ & 52.95$\pm$12.11 & \underline{56.55$\pm$3.70} & 61.53$\pm$3.03 & \underline{84.47$\pm$0.87} & 66.10$\pm$2.58  \\ [0.1cm]

  & TG-$\gnn$ & 52.96$\pm$6.81 & 56.52$\pm$2.85 & \underline{64.36$\pm$1.05} & 82.21$\pm$0.78 & \underline{66.24$\pm$1.11}  \\ [0.1cm]

  & CONCAT-$\gnn$ & \underline{54.54$\pm$1.64} & \textbf{56.63$\pm$1.68} & 62.58$\pm$1.45 & 81.80$\pm$0.77 & 65.83$\pm$0.04  \\ [0.1cm]

  & $\hact$-Net (Proposed) & \textbf{66.08$\pm$3.69} & 55.28$\pm$1.74 & \textbf{66.21$\pm$0.87} & \textbf{84.91$\pm$0.79} & \textbf{69.04$\pm$0.46} \\

  \bottomrule
\end{tabular}
\end{table*}

Comparing our proposed $\gnn$ solutions, we observe that $\cg$-$\gnn$ significantly outperforms CGC-Net, indicating the superiority of $\cnn$-based node feature initialization over handcrafted features, and the significance of $\gnn$s with expressive $\pna$ layers over Adaptive GraphSage in CGC-Net. We notice that $\cg$-$\gnn$ and $\tg$-$\gnn$ provide comparable performance overall. However, $\tg$-$\gnn$ performs better for Normal, Benign, and FEA subtypes, indicating the importance of tissue microenvironment information for these categories. In contrast, $\cg$-$\gnn$ surpasses $\tg$-$\gnn$ for UDH and ADH that rely on local nuclei-level information. For DCIS and Invasive, both $\cg$-$\gnn$ and $\tg$-$\gnn$ perform similarly, showing that both low-level and high-level information being useful for these categories. 
Further, both $\hact$-Net and $\concat$-$\gnn$ provide overall superior performance compared to all $\cnn$ and $\gnn$ baselines. $\hact$-Net significantly outperforms $\concat$-$\gnn$ indicating the significance of the proposed hierarchical modeling and learning. $\concat$-$\gnn$ produces overall comparable or superior performance to $\cg$-$\gnn$ and $\tg$-$\gnn$, although for individual classes, $\concat$-$\gnn$ is rarely better than the two, suggesting that it may be utilizing complementary information in $\cg$ and $\tg$ representations. Such complementary information is indeed best utilized by $\hact$-Net, with high per-class and the overall classification performance. Any of the proposed $\gnn$s often outperform all $\cnn$ baselines, establishing the potential of entity-based analysis in digital pathology.

\begin{figure}[!t]
\centering
\centerline{\includegraphics[width=1.05\linewidth]{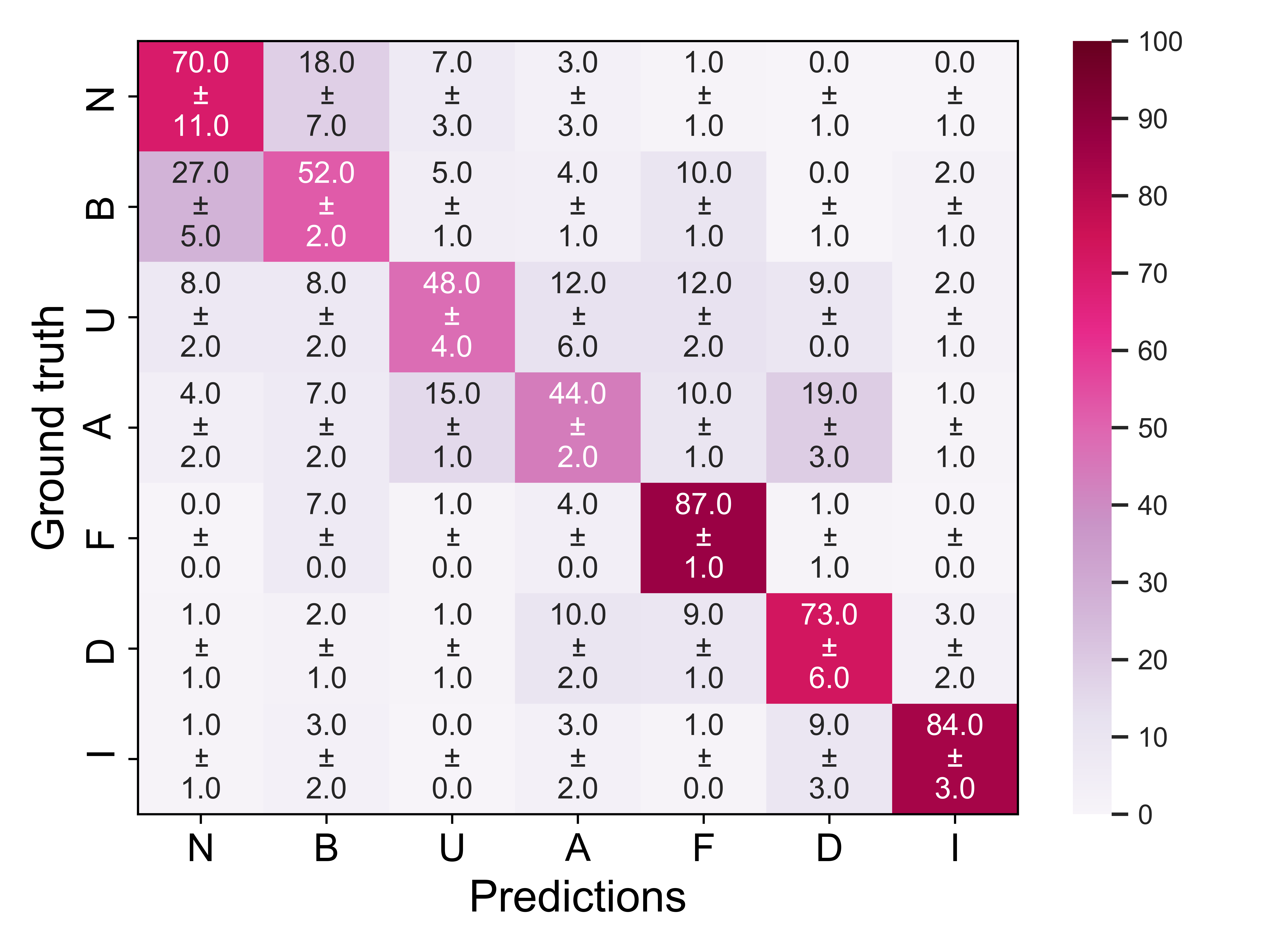}}
\caption{Mean and standard deviation of row-normalized 7-class confusion matrix for $\hact$-Net.}
\label{fig:confusion_matrix}
\end{figure}

Figure~\ref{fig:precision_recall} presents per-class precision and recall for $\cg$-$\gnn$, $\tg$-$\gnn$, $\concat$-$\gnn$, and $\hact$-Net. $\hact$-Net produces the highest precision values for most of the classes. The recall ranking between $\cg$-$\gnn$ and $\tg$-$\gnn$ varies across classes, whereas $\hact$-Net consistently yields good recall values. Further, standard deviation of class-wise precision and recall values are the lowest for $\hact$-Net, for most classes.
Figure~\ref{fig:confusion_matrix} presets row-normalized aggregated 7-class confusion matrix across three runs for $\hact$-Net. It indicates ambiguities between i)~Normal and Benign, ii)~UDH and ADH, and iii)~ADH and DCIS. 
Notably, these pair-wise classes bear high pathological ambiguity and are diagnostically very challenging.

\textbf{$\bullet$ Setting 2: 4-class classification:}
This setting categorizes $\troi$s into 4-classes based on cancer risk: Normal, Non-cancerous (Benign + UDH), Precancerous (ADH + FEA), and Cancerous (DCIS + Invasive). Classification performance of $\hact$-Net as well as $\cnn$ and $\gnn$ baselines are presented in Table~\ref{tab:4_class_results}. Single scale $\cnn$s exhibit the same behavior as in the 7-class setting. However, combining multiple magnifications in multi-scale $\cnn$s does not improve the classification performance over the single scale $\cnn$s. Among the baselines, CGC-Net and Patch-$\gnn$s perform comparable or inferior to the $\cnn$s, with a low-magnification $\cnn$(10$\times$) outperforming the others. 
Similarly to the 7-class setting, our proposed methods are superior to all the baselines. $\hact$-Net results in the best overall performance, with the best classification performance for Normal, Precancerous, and Cancerous categories. To highlight, $\hact$-Net achieves $\approx 66\%$ F1-score for the diagnostically challenging Precancerous category.

\begin{table*}[t]
\caption{Mean and standard deviation of weighted F1-scores for binary classification setting. Further, the aggregated mean and standard deviation for the six binary tasks are reported. Results are expressed in $\%$. The best result is in \textbf{bold} and the second best is \underline{underlined}.}
\label{tab:2_class_results}
\centering
\scriptsize 
\begin{tabular}{c|lcccccc|c}
  \toprule
  & Method & I vs & N+B+U vs & N vs  & B vs & A+F vs & A vs & Aggregated\\
  & & N+B+A+U+F+D & A+F+D & B+U  & U & D & F &\\
  \midrule
  \parbox[t]{2mm}{\multirow{5}{*}{\rotatebox[origin=c]{90}{$\cnn$}}} & $\cnn(10 \times)$ & 95.66$\pm$0.48 & 81.24$\pm$0.42 & 69.83$\pm$0.38 & 76.12$\pm$1.13 & 73.44$\pm$2.56 & 77.59$\pm$1.73 & 78.90$\pm$1.38\\ [0.1cm]

  & $\cnn(20 \times)$ & 92.39$\pm$0.37 & 80.84$\pm$0.36 & 66.52$\pm$2.14 & 74.75$\pm$1.51  & 67.87$\pm$1.82 & 71.78$\pm$2.53 & 75.69$\pm$1.68\\ [0.1cm]
 
  & $\cnn(40 \times)$ & 90.74$\pm$0.59 & 79.92$\pm$1.66 & 62.36$\pm$2.14 & 68.13$\pm$4.30  & 64.86$\pm$2.98 & 66.91$\pm$1.68 & 72.15$\pm$2.51\\ [0.1cm]

  & Multi-scale $\cnn (10\times+20\times)$ & 94.31$\pm$1.26 & 80.89$\pm$1.31 & 67.99$\pm$1.86 & 75.58$\pm$2.06  & 72.07$\pm$1.85 & 76.91$\pm$2.22 & 77.96$\pm$1.80\\ [0.1cm]

  & Multi-scale $\cnn (10\times+20\times+40\times)$ & 95.12$\pm$1.15 & 82.21$\pm$0.34 & 70.87$\pm$2.07 & 72.89$\pm$2.26  & 72.08$\pm$3.17 & 75.47$\pm$3.69 & 78.11$\pm$2.40\\ [0.1cm]
  
  \midrule
  \parbox[t]{2mm}{\multirow{4}{*}{\rotatebox[origin=c]{90}{$\gnn$}}} & CGG-Net & 91.60$\pm$2.09 & 79.73$\pm$1.53 & 63.67$\pm$3.12 & 62.37$\pm$3.00  & 81.56$\pm$1.56 & 73.80$\pm$5.41 & 75.46$\pm$3.09\\ [0.1cm]
  
  & Patch-$\gnn (10\times)$ & 95.80$\pm$0.43 & 76.53$\pm$0.32 & 72.57$\pm$1.10 & 72.87$\pm$3.07  & 77.17$\pm$0.85 & 78.26$\pm$2.60 & 78.87$\pm$1.75\\ [0.1cm]
  
  & Patch-$\gnn (20\times)$ & 93.70$\pm$0.36 & 76.63$\pm$1.40 & 70.10$\pm$1.90 & 69.77$\pm$3.13  & 74.10$\pm$0.10 & 81.03$\pm$1.85 & 77.55$\pm$1.78\\ [0.1cm]
  
  & Patch-$\gnn (40\times)$ & 92.40$\pm$0.95 & 74.43$\pm$0.64 & 71.10$\pm$1.74 & 67.40$\pm$2.46  & 72.97$\pm$0.66 & 76.40$\pm$1.95 & 75.78$\pm$1.56\\ [0.1cm]
  
  \midrule
  \parbox[t]{2mm}{\multirow{4}{*}{\rotatebox[origin=c]{90}{Ours}}} & CG-$\gnn$ (Ours) & 94.52$\pm$0.43 & \textbf{83.79$\pm$0.31} & \underline{75.71$\pm$1.68} & 73.15$\pm$3.32 & 77.48$\pm$1.68 & 84.33$\pm$0.54 & 81.50$\pm$1.70 \\ [0.1cm]
  
  & TG-$\gnn$ & \underline{96.00$\pm$0.56} & 80.38$\pm$0.80 & 69.51$\pm$3.12 & \underline{76.12$\pm$0.99} & \underline{80.67$\pm$0.22} & 84.18$\pm$3.56 & 81.14$\pm$2.02 \\ [0.1cm]
  
  & CONCAT-$\gnn$ & 95.91$\pm$0.56 & 83.21$\pm$0.68 & 71.84$\pm$1.46 & 75.67$\pm$1.81 & 80.14$\pm$2.60 & \underline{88.88$\pm$3.86} & \underline{82.61$\pm$2.15} \\ [0.1cm]
  
  & $\hact$-Net (Proposed) & \textbf{96.32$\pm$0.64} & \underline{83.63$\pm$0.73} & \textbf{76.84$\pm$0.68} & \textbf{77.66$\pm$0.37} & \textbf{81.11$\pm$0.72} & \textbf{89.35$\pm$0.26}  & \textbf{84.15$\pm$0.60}\\
  
  \bottomrule
\end{tabular}
\end{table*}

\begin{figure}[!t]
\centering
\centerline{\includegraphics[width=0.9\linewidth]{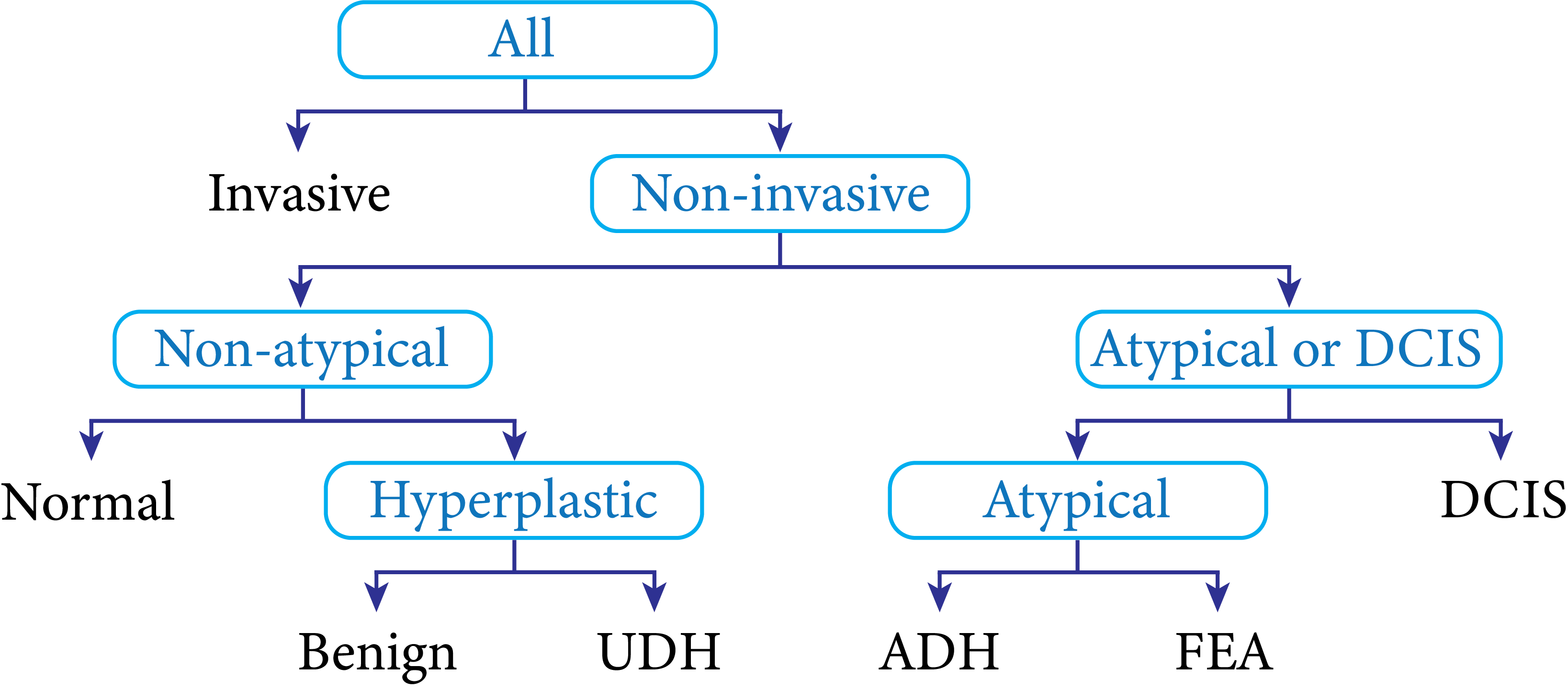}}
\caption{Decision tree used by pathologists for breast cancer diagnosis. The 7-class classification is simplified to a series of binary decision tasks, through which the diagnosis becomes more and more specific until the leaves, \ie the 7 diagnostic decision classes, are reached.}
\label{fig:decision_tree}
\end{figure}

\textbf{$\bullet$ Setting 3: Binary classifications:}
In this setting, we replicate the typical decision process of a pathologist following a diagnostic decision tree, presented in Figure~\ref{fig:decision_tree}. This is inspired by the classification scheme presented by~\cite{mercan18}, which mimics a pathologist's point-of-view for breast cancer subtyping. Note that such individual binary decisions are less constrained compared to a multi-class classification task, thereby allowing for better discrimination between a selected pair of classes as well as allowing one to incorporate risk in each such decision. Such binary classifiers can assist pathologists in categorizing ambiguous cases, at different parts of a decision tree. Table~\ref{tab:2_class_results} presents the results for six separate binary classification tasks, at the bifurcations in the decision tree. Results are consistent with the 7-class and 4-class classification settings, with $\hact$-Net consistently outperforming all baselines and providing the best aggregated score.

\begin{table*}[t]
\caption{Comparison between $\hact$-Net and domain expert pathologists for 7-class breast cancer subtyping on BRACS dataset. Per-class F1-scores, weighted F1-scores and accuracy for 7-class classification are presented. Results are expressed in $\%$. The best results are in \textbf{bold}.}
\label{tab:human_evaluation}
\centering
\scriptsize 
\begin{tabular}{lccccccc|c|c}
  \toprule
   & Normal & Benign & UDH & ADH & FEA & DCIS & Invasive & Weighted F1 & Weighted Accuracy \\
  \midrule
  Pathologist 1 & 67.53 & 53.92 & 41.90 & 36.00 & 19.13 & 71.59 & 94.00 & 55.30 & 56.71 \\ [0.1cm]
  Pathologist 2 & 47.83 & 52.94 & 25.00 & 35.37 & 65.22 & 68.00 & 94.00 & 57.07 & 57.99 \\ [0.1cm]
  Pathologist 3 & 39.66 & 49.59 & 49.43 & 42.29 & 54.12 & 65.19 & 89.47 & 56.71 & 56.55 \\ [0.1cm]
  \midrule
  Pathologist statistics & 51.57$\pm$11.70 & \textbf{52.15$\pm$1.85} & 38.78$\pm$10.22 & 37.89$\pm$3.12 & 46.16$\pm$19.64 & \textbf{68.26$\pm$2.62} & \textbf{92.49$\pm$2.14} & 56.36$\pm$0.76 & 57.08$\pm$0.64 \\ [0.1cm]
  $\hact$-Net statistics   & \textbf{61.56$\pm$2.15} & 47.49$\pm$2.94 & \textbf{43.60$\pm$1.86} & \textbf{40.42$\pm$2.55} & \textbf{74.22$\pm$1.41} & 66.44$\pm$2.57 & 88.40$\pm$0.19 & \textbf{61.53$\pm$0.87} & \textbf{63.21$\pm$0.27} \\
  \bottomrule
\end{tabular}
\end{table*}

\subsubsection{Domain expert comparison on BRACS dataset}
To further benchmark our proposed methodology as well as to assess the quality of the introduced BRACS dataset, we acquired annotations of the BRACS test set by additional independent pathologists. For such comparison with domain experts, we follow the evaluation protocol in~\cite{elmore15}.
We recruited three board-certified pathologists (other than the original three pathologists who provided the initial annotations, namely our ground truth labels), from three different medical centers, to further ensure independence: $\bullet$~National Cancer Institute- IRCCS-Fondazione Pascale, Naples, Italy; $\bullet$~Lausanne University Hospital, CHUV, Lausanne, Switzerland; and $\bullet$~Aurigen, Centre de Pathologie, Lausanne, Switzerland. 
These experts are specialized in breast pathology and have been in practice for over twenty years. The pathologists independently and remotely annotated BRACS test set $\troi$s, without having access to respective WSIs. This protocol ensures equal field-of-view for all the pathologists as well as our methodology. 

These new pathologist annotations are compared to the ground truth labels, with the results shown in Table~\ref{tab:human_evaluation}. We present per-class F1-scores, overall weighted F1-score, and overall weighted accuracy for each individual pathologist. Further, we include the aggregated statistics of all three pathologists for benchmarking $\hact$-Net with domain experts. Table~\ref{tab:human_evaluation} indicates that $\hact$-Net outperforms the domain experts in distinguishing $\troi$s of diagnostically challenging classes, \ie atypia and hyperplasias, while yielding comparable performance for the normal and cancerous categories. 
Per-class standard deviations of pathologists' statistics illustrate the expected high inter-observer variability in breast cancer diagnosis.
Compared to the pathologists, $\hact$-Net yields a superior weighted accuracy and weighted F1 given the ground truth diagnoses for the 7-class classification task. 

\begin{table}
\caption{Concordance among three independent pathologists for annotating BRACS test dataset. Results are expressed in $\%$.}
\label{tab:human_concordance}
\centering
\scriptsize 
\begin{tabular}{lcccc}
  \toprule
   & Pathologist 1 & Pathologist 2 & Pathologist 3 & Ground truth\\
  \midrule
  Pathologist 1 & - & 47.60 & 50.96 & 56.71 \\ [0.1cm]
  Pathologist 2 & - & - & 64.38 & 57.99 \\ [0.1cm]
  Pathologist 3 & - & - & - & 56.55 \\ [0.1cm]
  \bottomrule
\end{tabular}
\end{table}

To benchmark the BRACS dataset with respect to the dataset by~\cite{elmore15}, we compare the aggregated pathologist statistics on both datasets for the same set of classes, \ie Benign without atypia (Normal + Benign + UDH), Atypia (ADH + FEA), DCIS, and Invasive. Note that the dataset by~\cite{elmore15} consists of 240 breast biopsy slides, while BRACS consists of 626 $\troi$ images. For the dataset by~\cite{elmore15}, class-wise concordance rates (class-weighted average accuracy of 115 pathologists to a three-expert consensus) are 87\%, 48\%, 84\%, and 96\%, respectively for the four aforementioned classes. For BRACS, the similar class-wise concordance rates are 87\%, 50\%, 72\%, and 90\%, respectively. The class-wise concordance rates exhibit a similar trend in both datasets. Differences can be attributed to differing fields-of-view, \ie $\troi$ vs.\ WSI, accessible to the pathologist during annotation. 

Table~\ref{tab:human_concordance} presents the inter-observer concordance rates for the BRACS test set. We notice significant differences in the concordance rates between pathologists 2\,vs.\.3 and pathologist 1\,vs.\,the other two. This can be reasoned to differing histopathology practices across different regions. 

\begin{figure*}
\centering
\centerline{\includegraphics[width=0.98\textwidth]{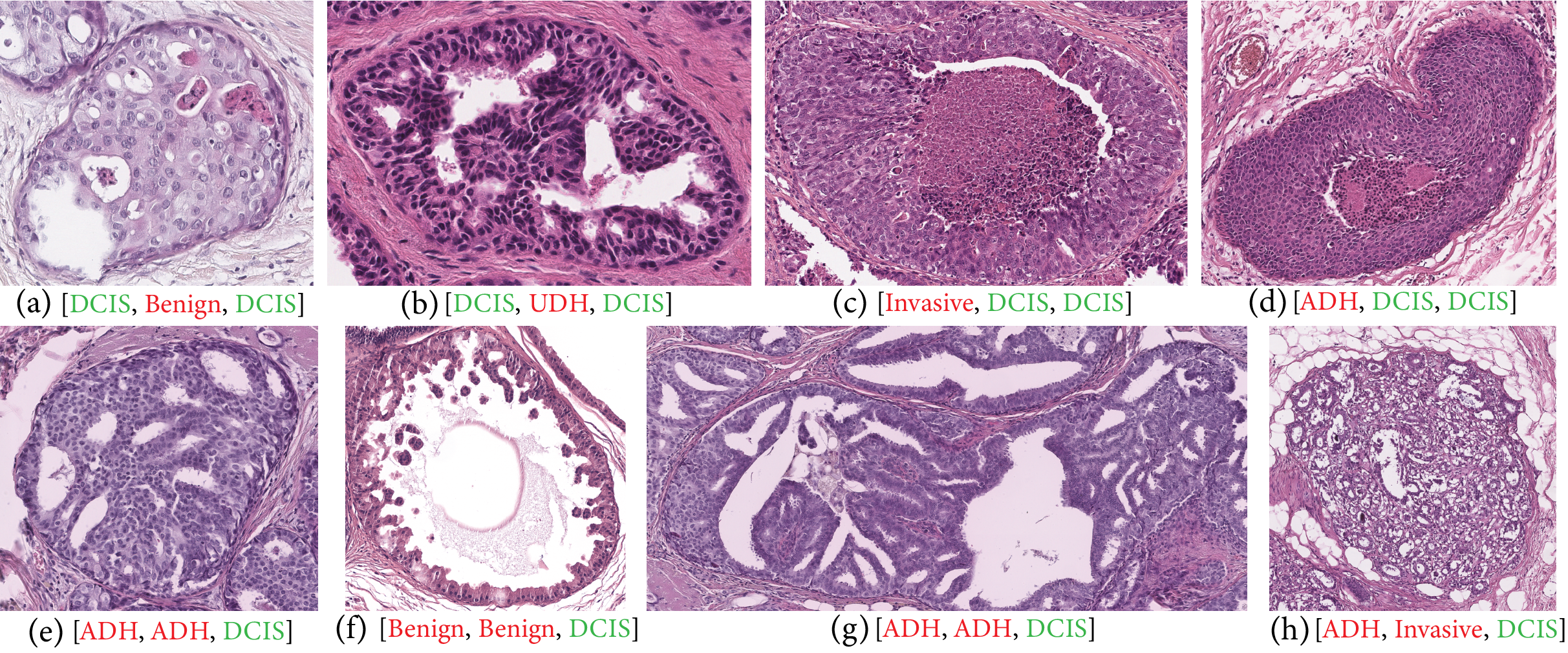}}
\caption{Qualitative comparison of $\cg$-$\gnn$, $\tg$-$\gnn$, and $\hact$-Net for 7-class classification. Predictions by the classifiers are noted below each example. \textcolor{red}{Red} and \textcolor{green}{Green} denote incorrect and correct classification, respectively. 
(a,b)~$\troi$s which $\tg$-$\gnn$ misclassifies, while $\cg$-$\gnn$ and $\hact$-Net classify correctly by using the nuclei characteristics. 
(c,d)~$\troi$s misclassified by $\cg$-$\gnn$, while correctly classified by $\tg$-$\gnn$ and $\hact$-Net by using context information from necrotic regions. 
(e,f,g,h)~$\troi$s which both $\cg$-$\gnn$ and $\tg$-$\gnn$ misclassify, where $\hact$-Net classifies correctly by utilizing both cell and tissue microenvironments together. (Figure is best viewed in color)}
\label{fig:qualitative}
\end{figure*}

\begin{figure*}[!t]
\centering
\centerline{\includegraphics[width=0.98\textwidth]{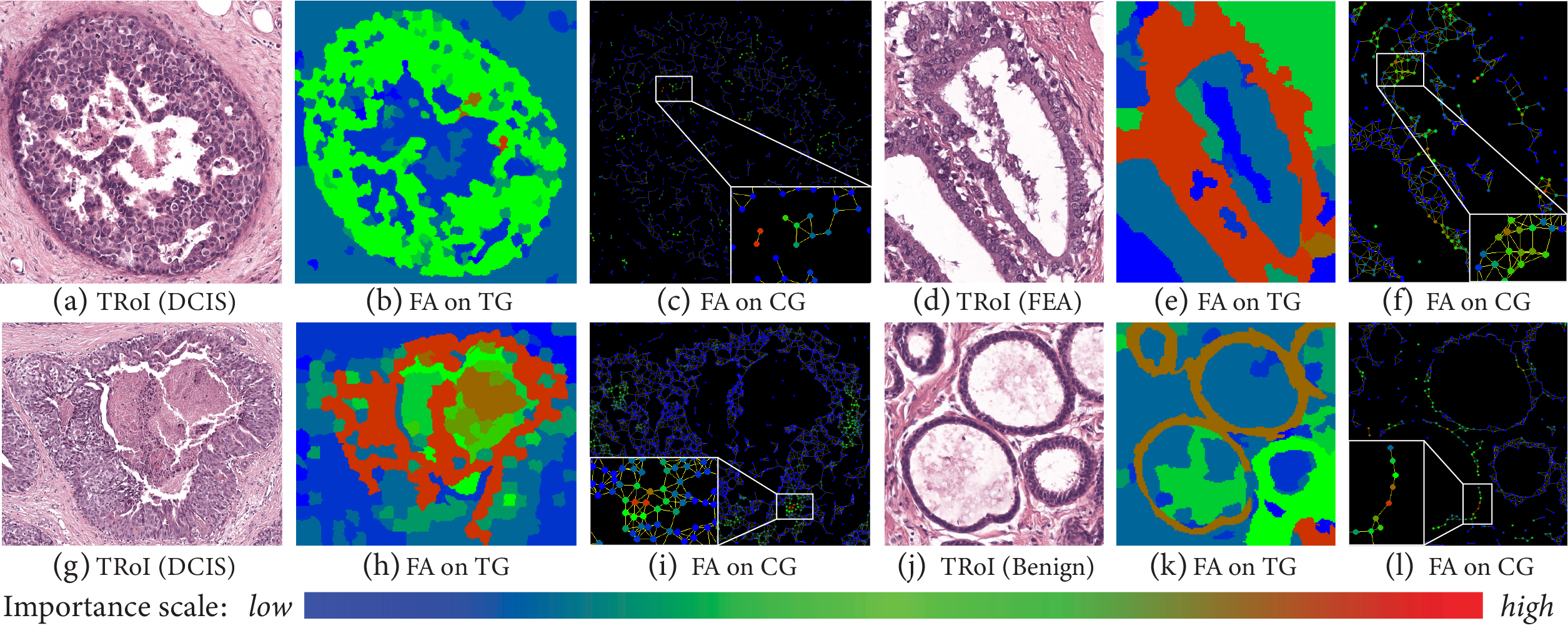}}
\caption{Feature attribution (FA) maps of $\hact$-Net on $\tg$ and $\cg$ for four sample $\troi$s for 7-class classification:
Sample $\troi$s of (a,g)~DCIS, (d)~FEA, and (j)~Benign classes, with their corresponding feature attribution maps on (b,h,e,k)~$\tg$ and (c,i,f,l)~$\cg$. (Figure is best viewed in color)}
\label{fig:qualitative_gradcam}
\end{figure*}

\subsubsection{Classification results on BACH dataset}
We evaluate the methods on the public BACH dataset. Considering its smaller training set of 400 images, we employ different image augmentation techniques for training $\hact$-Net. To this end, we employ rotation, mirroring, and color augmentations on the training images before extracting $\hact$ graph representations. We do not use other graph augmentation techniques, such as random node and edge dropping, since these augmentations may hamper the meaningful topological distribution of the biological entities. The implementation strategies and hyperparameters in Section~\ref{sec:implementation} are employed for training $\hact$-Net.
Classification performance of $\hact$-Net and the current state-of-the-art results on the BACH dataset are listed in Table~\ref{tab:bach_results}. Our predictions have been evaluated independently by the organizers of the BACH challenge, ensuring a fair with the other methods.
$\hact$-Net results in comparable classification accuracy with the state-of-the-art methods. The difference in the accuracies are not significant considering only 100 $\troi$s in the test set.
Notably, our methodology employs a single, unified network, where the other listed competitors all employing an ensemble strategy with multiple networks at inference.

\begin{table}
\caption{Accuracy of 4-class breast cancer subtyping in BACH dataset. Results are expressed in $\%$.}
\label{tab:bach_results}
\centering
\scriptsize 
\begin{tabular}{p{0.11\textwidth}|l|c}
  \toprule
  & Methods & Accuracy \\
  \midrule
  \parbox[t]{10mm}{\multirow{6}{*}{\shortstack{Ensemble \\ networks \\ \cite{bach18} \\ \cite{aresta19}}}} &
  \textcolor{blue}{Wang et al. (2019)} & 95.00 \\ [0.1cm]
  & \cite{marami18} & 94.00 \\ [0.1cm]
  & \textcolor{blue}{Yang et al. (2019)} & 93.00 \\ [0.1cm]
  & \cite{chennamsetty18} & 87.00 \\ [0.1cm]
  & \textcolor{blue}{Kwok et al. (2018)} & 87.00 \\ [0.1cm]
  & \cite{brancati18} & 86.00 \\ [0.1cm]
  \midrule
  Single network & $\hact$-Net & 91.00 \\ [0.1cm]
  \bottomrule
\end{tabular}
\end{table}

\subsection{Qualitative analysis}
Qualitative assessment of some results with our proposed method on the BRACS dataset is presented in Figure~\ref{fig:qualitative}, where we compare our method with standalone $\cg$ and $\tg$ based learning.
In Figure~\ref{fig:qualitative_gradcam}, we employ \textsc{GraphGradCAM}~\cite{pope19, jaume21}, a post-hoc gradient based feature attribution technique, to highlight the nuclei and tissue-region nodes in $\cg$ and $\tg$, respectively, to show what $\hact$-Net focuses on while classifying the $\troi$s.  
Given the DCIS examples in Figs.~\ref{fig:qualitative_gradcam}\textcolor{blue}{(a-c\&g-i)}, $\hact$-Net is seen to focus on the diagnostically relevant tumorous epithelial and the necrotic regions in $\tg$, while ignoring the less important stroma and lumen, cf.\ Figs.~\ref{fig:qualitative_gradcam}\textcolor{blue}{(b,h)}.
Further, within these relevant tissue regions, $\hact$-Net focuses on a subset of tumorous epithelial nuclei in $\cg$, as seen in Figs.~\ref{fig:qualitative_gradcam}\textcolor{blue}{(c,i)}.
Interestingly, we observe in Figs.~\ref{fig:qualitative_gradcam}\textcolor{blue}{(h,i)} that $\hact$-Net utilizes complementary information from the necrotic region captured by $\tg$, but not by $\cg$.
Similar observations of $\hact$-Net considering the diagnostically relevant regions can also be made for FEA and Benign examples seen in Figs~\ref{fig:qualitative_gradcam}\textcolor{blue}{(d-f\&j-l)}. 
Noticeably, such feature attribution analysis $\gnn$s localizes and highlights the focus of deep networks in the given entity-paradigm, which is both more interpretable and more explainable compared to feature attribution strategies in a pixel-paradigm~\cite{jaume20,jaume21}.


\section{Conclusion}
\label{sec:conclusion}
Pixel-based processing of pathology images suffers from the context-resolution trade-off, and misses the notion of biological entity and tissue composition. In this work, we propose an entity-based tissue representation and learning to address the issues with pixel-based processing.
To that end, our two specific contributions are:
(i)~a hierarchical entity-graph representation of a tissue image by incorporating multisets of pathologically intuitive biological entities, and
(ii)~a hierarchical graph neural network for sequentially processing the entity-graph representation for mapping tissue compositions to respective tissue classes.
Further, we introduce BReAst Cancer Subtyping (BRACS), a large cohort of breast tumor regions-of-interest, annotated with breast cancer subtypes. BRACS encompasses seven breast cancer subtypes to represent a realistic breast cancer diagnosis scenario.
Using BRACS as well as a public breast cancer subtyping dataset BACH, we demonstrate herein the superior performance of our proposed methodology for classifying breast tumor regions-of-interest into cancer subtypes.
Under various experimental settings, our methodology is shown to outperform state-of-the-art pixel-based and entity-graph based classification approaches.
Furthermore, we benchmark our methodology on the BRACS dataset by comparing it to three independent pathologists. Notably, our method achieves better performance for per-cancer subtype and overall aggregated classification. 
Although we have evaluated our method for breast cancer classification, the technology is easily extendable to other tissue types and diseases. Notably, the proposed hierarchical graph methodology can also be adapted to other image modalities, such as natural images, multiplexed images, hyperspectral images, satellite images, and other medical imaging domains, by utilizing domain and task-specific entities.
\\

\bibliography{main}
\bibliographystyle{IEEEtran}

\end{document}